\documentclass[11pt]{article}
\usepackage[utf8]{inputenc} 
\usepackage[T1]{fontenc}    
\usepackage{hyperref}       
\usepackage{url}            
\usepackage{booktabs}       
\usepackage{amsfonts}       
\usepackage{nicefrac}       
\usepackage{microtype}      
\usepackage{xcolor}         
\usepackage[pdftex]{graphicx}
\usepackage{amsmath,amssymb,subfigure,epsfig,bbm,stmaryrd,MnSymbol,mathrsfs,url}
\usepackage{pstricks,framed,tcolorbox}
\DeclareGraphicsExtensions{.pdf}
\usepackage{amstext}
\usepackage{array,xcolor}
\usepackage{color,soul}
\usepackage{hyperref}
\usepackage{blkarray}
\usepackage{amsmath}
\usepackage{mdframed}

\usepackage{amsmath}
\usepackage[all,cmtip]{xy}
\usepackage{chemfig}
\usepackage{booktabs,makecell}
\usepackage{diagbox}
\usepackage[T1]{fontenc}
\usepackage[utf8]{inputenc}
\usepackage[font=small,labelfont=bf,tableposition=top]{caption}
\usepackage{easybmat}
\usepackage{multirow,bigdelim}
\usepackage{makeidx,overpic,tcolorbox}
\makeindex
\usepackage{extarrows} 
\usepackage{graphicx}

\usepackage{amsthm}
\newtheorem{theorem}{Theorem}
\newtheorem{lemma}{Lemma}

\newtheorem{proposition}{Proposition}

\newcommand{\R}{\mathbb{R}}

\usepackage{xr}
\makeatletter
\newcommand*{\addFileDependency}[1]{
  \typeout{(#1)}
  \@addtofilelist{#1}
  \IfFileExists{#1}{}{\typeout{No file #1.}}
}
\makeatother

\newcommand{\EE}{\mathbb{E}}
\newcommand{\balpha}{{\boldsymbol{\alpha}}}
\newcommand{\bbeta}{\boldsymbol{\beta}}

\newcommand{\bdelta}{\boldsymbol{\delta}}
\newcommand{\bmu}{\boldsymbol{\mu}}
\newcommand{\bxi}{\boldsymbol{\xi}}
\newcommand{\bSigma}{\boldsymbol{\Sigma}}
\newcommand{\bsigma}{\boldsymbol{\sigma}}

\newcommand{\Mi}{{\mathcal{M}_i}}
\newcommand{\Ai}{{\mathcal{A}_i}}

\newcommand{\AAA}{{\mathcal{AA}}}
\newcommand{\AM}{{\mathcal{AM}}}
\newcommand{\MA}{{\mathcal{MA}}}
\newcommand{\MM}{{\mathcal{MM}}}
\newcommand{\prox}{{\textup{\texttt{prox}}}}
\newcommand{\eM}{{\mathcal{M}}}

\newcommand{\vct}[1]{\boldsymbol{#1}}
\newcommand{\mtx}[1]{\boldsymbol{#1}}

\newcommand{\trace}{\operatorname{trace}}

\DeclareMathOperator*{\minimize}{\text{minimize}}

\DeclareMathOperator*{\argmin}{arg\,min}
\DeclareMathOperator*{\argmax}{arg\,max}

\newcommand{\va}{\vct{a}}
\newcommand{\vb}{\vct{b}}

\newcommand{\vd}{\vct{d}}

\newcommand{\vq}{\vct{q}}

\newcommand{\vu}{\vct{u}}
\newcommand{\vv}{\vct{v}}

\newcommand{\vx}{\vct{x}}
\newcommand{\vy}{\vct{y}}
\newcommand{\vz}{\vct{z}}
\newcommand{\mA}{\mtx{A}}
\newcommand{\mB}{\mtx{B}}
\newcommand{\mC}{\mtx{C}}
\newcommand{\mD}{\mtx{D}}

\newcommand{\mI}{\mtx{I}}

\newcommand{\mQ}{\mtx{Q}}

\newcommand{\mS}{\mtx{S}}

\newcommand{\mU}{\mtx{U}}
\newcommand{\mV}{\mtx{V}}

\newcommand{\mX}{\mtx{X}}

\DeclareMathOperator{\sign}{sign}

\usepackage{tikz,siunitx}
\usetikzlibrary{arrows}
\usetikzlibrary{matrix,positioning,calc}
\usetikzlibrary{decorations.pathmorphing}
\tikzset{snake it/.style={-stealth,
		decoration={snake, 
			amplitude = 1.5mm,
			segment length = 2.5mm,
			post length=2.9mm},decorate}}

\usepackage{hyperref}

\date{}

\title{\textbf{RIGID: Robust Linear Regression with Missing Data}}
\author{%
  Alireza Aghasi\thanks{Department of Electrical Engineering and Computer Science, Email: \texttt{alireza.aghasi@oregonstate.edu} (Corresponding Author)}~,
  Mohammad Javad Feizollahi\thanks{Department of Data Science and Analytics, Georgia State University, Atlanta, GA}, and  
  Saeed Ghadimi\thanks{Department of Management Sciences,
  University of Waterloo, ON, Canada}\\

}

\begin{document}
\maketitle

\begin{abstract}
We present a robust framework to perform linear regression with missing entries in the features. By considering an elliptical data distribution, and specifically a multivariate normal model, we are able to conditionally formulate a distribution for the missing entries and present a robust framework, which minimizes the worst case error caused by the uncertainty about the missing data. We show that the proposed formulation, which naturally takes into account the dependency between different variables, ultimately reduces to a convex program, for  which a customized and scalable solver can be delivered. In addition to a detailed analysis to deliver such solver, we also asymptoticly analyze the behavior of the proposed framework, and present technical discussions to estimate the required input parameters. We complement our analysis with experiments performed on synthetic, semi-synthetic, and real data, and show how the proposed formulation improves the prediction accuracy and robustness, and outperforms the competing techniques.

Missing data is a common problem associated with many datasets in machine learning. With the significant increase in using robust optimization techniques to train machine learning models, this paper presents a novel robust regression framework that operates by minimizing the uncertainty associated with missing data. The proposed approach allows training models with incomplete data, while minimizing the impact of uncertainty associated with the unavailable data.  The ideas developed in this paper can be generalized beyond linear models and elliptical data distributions. 
\end{abstract}

\section{Introduction}

Missing data occurs in many data-related disciplines, such as economics, social sciences, health care, marketing and entertainment industry. Various sources could be the cause of missing data, such as different standards and protocols to collect the data (e.g., in healthcare), confidentiality of the data (e.g., in social sciences), and limitations in installing measuring sensors (e.g., in empirical physical models). Rubin \cite{Rubin76} classifies the missing data mechanisms into three main categories: (1) missing completely at random (MCAR), where the probability of missing is the same for all samples; (2) missing at random (MAR), where the probability of missing is the same within groups of the samples; (3) missing not at random (MNAR), covering other mechanisms which are neither MCAR, nor MAR.

The common practice to deal with missing data is to use imputation from the observed portion of the data. Especially in the context of supervised learning problems, the construction of predictive models commonly follow two steps, first, imputing the data, then, training predictive models which operate with full data. Great effort has been devoted to providing good imputation from the observed data ~\cite{BuGr11,YuYao17}. Once the completed dataset has been recovered with imputation, different predictive models have been exploited for the out-of-sample prediction.
The simplest and probably the most widely-used imputation method is to impute with a constant, namely, the mean of the observed data.  More robust and sophisticated techniques such as maximum likelihood estimation, Bayes and multiple imputation, and iterative predictive models are also considered (see~\cite{LitRub19} and the references therein). Following the imputation, different statistical models have been used for the prediction (e.g.,~\cite{josse2020consistency,ElGe19,fan2019precision}). Recently, more attention has been given to concurrent imputation and prediction techniques ~\cite{bertsimas2021prediction,chandrasekher2020imputation,morvan2021whats}, which seem to present more promising results. This is, especially something that has been discussed in \cite{morvan2021whats}.  

On the other hand, over the past few years, there has been an increasing interest towards robust predictive models, especially, those designed to cope with adversarial attacks \cite{madry2018towards}. While, robust regression techniques have been extensively analyzed for linear models (see for example \cite{YuYao17} and the references therein), this problem is not well-studied under the presence of missing data. In \cite{Han14}, the authors propose a multiply robust estimation that allows multiple models for the missing mechanism and the data distribution. Our focus in this paper is to present a robust framework for linear regression problems, when the data are missing. In fact, robustness is used to minimize the uncertainty that negatively affects the model training process. While, our proposed approach trains a model through a single, unified optimization program, similar to \cite{chandrasekher2020imputation} one may view it as a two stage process of first imputing the data with the conditional mean, and then performing a robust regression that uses the missing data conditional statistics.

Our contribution in this paper consists of the following aspects. First, we show that robust linear regression with missing data can be reduced from a min-max program to an unconstrained convex program. The resulting program admits a special structure which further enables us to reformulate it as minimizing a convex function subject to linear constraints, for which we can obtain a closed-form solution to the corresponding proximal operator. As a result, we can design an alternating direction method of multipliers (ADMM) framework to obtain the robust solution.
Second, by considering a reasonably realistic setup, we present a closed-form representation of the expected robust risk as a proxy for big data regimes. Despite its complex structure, we present some characteristics of the robust risk and its solution. We also discuss the process of estimating the parameters involved in our robust formulation, and present mathematical guarantees for their estimation quality. 

The rest of the paper is organized as follows. In Section~\ref{PS}, we present the statement of our problem and reformulate the robust regression of incomplete data as a convex optimization problem. In Section~\ref{cvx}, we discuss an ADMM approach to solve the convex reformulated problem. In Section~\ref{robust}, we focus on the asymptotic behavior of our loss function, and its characterization.  Section~\ref{sec:EstSufSt} discusses the estimation of the data mean and the covariance matrix from incomplete data, to be used in our robust formulation. Finally, we present our numerical experiments in Section~\ref{numerical} followed by some concluding remarks in Section \ref{conclude}. The proofs of the technical results are mainly presented in the Appendix, however, to abide by the page limit, some proofs, technical discussions, and experimental details are presented as Supplementary Material. 

\paragraph{Notation.} 
We follow standard notations. Vectors and matrices are written with boldface. Unless specified, for a given index set $A$, $\vx_A$ denotes the restriction of the vector $\vx$ to the indices in $A$. Similarly, $\mX_{AA'}$ represents the restriction of the matrix $\mX$ to rows in $A$, and columns in $A'$. The complement of the set $A$ is denoted by $A^c$. Given an integer $p$, the set $\{1,2,\ldots,p\}$ is denoted by $[p]$. Given a matrix $\mS$, the notations $\mS\succ \boldsymbol{0}$ and $\mS\succeq\boldsymbol{0}$ indicate $\mS$ being positive definite, and positive semi-definite, respectively.  

\section{Problem Statement}\label{PS}
Consider the linear regression model $y = \vx^\top \bbeta_0 + \epsilon$,    
where $\vx\in\R^p$ is the feature vector, and $\epsilon\!\sim\! \mathcal{N}(0,\sigma^2)$ is an independent noise term (to avoid unnecessary notation complications, the intercept term for the model is absorbed as a constant feature in $\vx$). Throughout the paper we assume that $\vx$ follows a multivariate normal distribution as $\vx\sim\mathcal{N}(\bmu,\bSigma)$. In fact, the proposed technique is generally applicable to cases where $\vx$ follows an elliptical distribution, such as normal, $t$, Laplace and logistic distributions, however, to have closed-form conditional expressions the derivations are done for the normal case.

Having $n$ independent realizations of the model as $(\vx_i, y_i)$, $i\in[n]$, a typical fitting problem to estimate $\bbeta_0$ consists of minimizing the empirical risk: 
\begin{equation}\label{eq:riskminfull}
  \minimize_{\bbeta}~~ \frac{1}{2n}\sum_{i=1}^n \left (y_i - \vx_i^\top \bbeta\right)^2.  
\end{equation}
The problem of interest in this paper is to address this fitting problem when possibly a subset of the elements in each sample $\vx_i$ is missing. We are specifically interested in devising a statistically robust and consistent way to perform this task.  

Towards this goal, we consider a similar problem setup, but now for the sample $\vx_i$ the entries corresponding to the index set $\Mi\subseteq[p]$ are missing, and only a subset of the elements indexed by $\Ai=\Mi^c$ are available. To avoid disrupting the formulation flow, for now we assume that the mean and covariance of $\vx_i$, $\bmu$ and $\bSigma$, are known. Later in Section \ref{sec:EstSufSt}, we provide discussions and mathematical guarantees to estimate these quantities in missing data regimes.

In the aforementioned missing data regime, while the $\vx_{i,\Mi}$ portion of $\vx_i$ is completely unobserved, still a statistical characterization of its distribution is possible conditioned on the observed potion. As a standard result in multivariate analysis (e.g., see \S 1.2 of \cite{muirhead2009aspects}), we know that when $\vx_i\!\sim\!\mathcal{N}(\bmu,\bSigma)$, and $\bSigma_{\Ai\Ai}\succ \boldsymbol{0}$,  the distribution of $\vx_{i,\Mi}$ conditioned on the observation of $\vx_{i,\Ai}$ also follows a multivariate normal distribution as $\vx_{i,\Mi}\!\sim\! \mathcal{N}(\bar\bmu_i, \bar\bSigma_i)$, where  
\vspace{-.1cm} 
\begin{equation*}\label{eq:xbmu}
\bar \bmu_i=\bmu_{\mathcal{M}_i}+  \bSigma_{\mathcal{M}_i \mathcal{A}_i} \bSigma_{\mathcal{A}_i \mathcal{A}_i}^{-1} (\vx_{i,\mathcal{A}_i}-\bmu_{\mathcal{A}_i}),
~~~
\bar \bSigma_i=   \bSigma_{ \mathcal{M}_i \mathcal{M}_i}-  \bSigma_{\mathcal{M}_i \mathcal{A}_i} \bSigma_{\mathcal{A}_i \mathcal{A}_i}^{-1} \bSigma_{\mathcal{A}_i \mathcal{M}_i} \triangleq  \bSigma \slash \bSigma_{\Ai\Ai}.\vspace{-.1cm} 
\end{equation*}
The expression $\bSigma \slash \bSigma_{\Ai\Ai}$ denotes the Schur complement of $\bSigma_{\Ai\Ai}$ in $\bSigma$. Consequently, when $\bbeta$ is known, the random variable $\vx_{i,\mathcal{M}_i}^\top \bbeta_{\mathcal{M}_i}$ follows the normal distribution $\mathcal{N}(\bar \bmu_i^\top  \bbeta_{\mathcal{M}_i}  ,\| \bbeta_{\mathcal{M}_i}\|_{\bar \bSigma_i}^2 )$, where $\| \bbeta_{\mathcal{M}_i}\|_{\bar \bSigma_i}^2 = \bbeta_{\mathcal{M}_i}^\top \bar \bSigma_i \bbeta_{\mathcal{M}_i}$. Because of the normality, this random variable maximally concentrates around its mean, and one can characterize the concentration region as
\begin{equation*}\label{eq:concentrationRegion}
\mathcal{R}_i(\bbeta;\gamma ) \triangleq \left\{\vz:~\! \left |\vz^\top \bbeta_{\mathcal{M}_i} - \bar\bmu_i^\top \bbeta_{\mathcal{M}_i} \right | \leq \gamma \left\| \bbeta_{\mathcal{M}_i}\right\|_{\bar \bSigma_i} \right\},
\end{equation*} 
where $\gamma\geq 0$ controls the deviation from the mean. Since $\vx_i^\top\bbeta = \vx_{i,\Ai}^\top \bbeta_\Ai + \vx_{i,\Mi}^\top \bbeta_\Mi$, and such concentration domain is available for the unobserved portion of $\vx_i$, we can introduce a robust counterpart of \eqref{eq:riskminfull} as
\begin{equation}\label{eq:riskminRobust}
  \minimize_{\bbeta} \!\! \max_{\substack{\vx_{i,\Mi}\in\mathcal{R}_i(\bbeta;\gamma)\\ i\in [n]}} \frac{1}{2n}\!\sum_{i=1}^n \!\left ( y_i\! -\! \vx_{i,\Ai}^\top \bbeta_\Ai\! -\! \vx_{i,\Mi}^\top \bbeta_\Mi\!\right)^2\!\!, 
\end{equation}
where we try to minimize the worst case error caused by the uncertainty about the value of $\vx_{i,\Mi}$. The min-max program in \eqref{eq:riskminRobust} is central to our robust linear regression in the missing data (RIGID) framework, where $\gamma$ serves as a free parameter. As will be noted in the sequel, this program can be further simplified and cast as a convex program. 

Considering the functions 
\begin{equation}\label{eq:defL}
\mathcal{L}_i(\bbeta) \triangleq  \!\!\! \max_{\vx_{i,\Mi}\!\!\in\mathcal{R}_i(\bbeta;\gamma)} \frac{1}{2}\left ( y_i\! -\! \vx_{i,\Ai}^\top \bbeta_\Ai\! -\! \vx_{i,\Mi}^\top \bbeta_\Mi\!\right)^2\!\!,
\end{equation}
by the interchangeability of the max and sum operations (thanks to the decoupling of the inner loss), program \eqref{eq:riskminRobust} can be cast as
\begin{equation}\label{eq:minL}
\minimize_{\bbeta} ~~\frac{1}{n}\sum_{i=1}^n \mathcal{L}_i(\bbeta). 
\end{equation}
From a convex analysis standpoint, if for a given $\vx_{i,\Mi}$ the function and constraint set in \eqref{eq:defL} were both convex in $\bbeta$, one could claim that  $\mathcal{L}_i(\bbeta)$ is a supremum over a family of convex functions and hence would itself be convex (\S 3.2.3 \cite{boyd2004convex}). While the constraint set is not convex in $\bbeta$, we can still present a closed-form convex form of $\mathcal{L}_i(\bbeta)$ as follows.
\begin{proposition}\label{prop:Li}
    Consider the function $\mathcal{L}_i(\bbeta)$ defined in \eqref{eq:defL} with $\bar\bSigma_i\succeq \boldsymbol{0}$ and $\gamma\geq 0$ (elements contributing to the constraint set $\mathcal{R}_i(\bbeta;\gamma)$). Then 
    \begin{equation}\label{eq:defL2}
    \mathcal{L}_i(\bbeta) =\frac{1}{2}\left (\left | y_i\! -\! \vx_{i,\Ai}^\top \bbeta_\Ai\! -\! \bar\bmu_i^\top \bbeta_\Mi\!\right | +  \gamma \left\| \bbeta_{\mathcal{M}_i}\right\|_{\bar \bSigma_i} \right)^2\!\!,
\end{equation}
and it is a convex function of $\bbeta$. 
\end{proposition}
\begin{proof}
See \S~\ref{prop1Prf} of the Supplementary Material. 
\end{proof}
Based on this result, the RIGID problem in \eqref{eq:riskminRobust} ultimately reduces to the convex program \eqref{eq:minL}, where $\mathcal{L}_i(\bbeta)$ follows the formulation in \eqref{eq:defL2}. The parameter $\gamma$ is a free parameter to tune the model and can be chosen through cross-validation. In the next section, we will discuss some computational tools to solve this convex optimization problem. 

\section{The Solution Method for RIGID}\label{cvx}
The focus of this section is developing a computational algorithm to solve problem \eqref{eq:minL}, when $\mathcal{L}_i(\bbeta)$ follows the formulation in \eqref{eq:defL2}. To this end, we propose an alternating direction method of multipliers (ADMM) scheme, which is computationally efficient and scalable when a closed-form proximal operator for the problem is available. Given a closed proper function $\phi:\R^m\to\R\cup\{+\infty\}$, the effective domain of $\phi$ is specified as $\texttt{dom}(\phi) = \{\vz \in \R^m: \phi(\vz)<+\infty \}$. For this function, the proximal operator is defined as 
$\prox_{\phi(.)}(\vz') = \argmin_{\vz}  ~ \phi(\vz) + \frac{1}{2}\|\vz - \vz'\|^2.$
Our proposed ADMM scheme exploits a proximal operator to carry out the RIGID convex program. As detailed in Theorem \ref{mainProxTh} at the end of this section, a careful derivation allows accessing such closed-form expression.

Given $\gamma\geq 0$, RIGID can be viewed as an instance of the following convex program
 \begin{align}\label{convProgGen}
\minimize_{\bbeta~\!\in~\!\R^p} ~~ \frac{1}{2n}\sum_{i=1}^n ~ \left (\left | a_i - \vb_i^\top\bbeta \right | +  \gamma \left\|\mC_i^\top \bbeta \right\|\right)^2. 
\end{align}
To see the connection, consider $\mS_i\in\R^{|\Mi|\times p}$ as the selection matrix which returns $\bbeta_\Mi$ when applied to $\bbeta$, as $\bbeta_\Mi = \mS_i\bbeta$. Such matrix can be easily obtained by restricting the rows of an identity matrix of size $p$ to $\Mi$. One can then immediately see that RIGID is an instance of \eqref{convProgGen} when $a_i$ is picked to be $y_i$, $\vb_i$ is set to be the concatenation of the vectors $\vx_{i,\Ai}$ and $\bar \bmu_i$ (in the right index order), and $\mC_i$ is taken to be $\mS_i^\top \bar\mC_i$, where $\bar\mC_i$ is the Cholesky factor of $\bar \bSigma_i$ (i.e., $\bar\bSigma_i = \bar\mC_i\bar\mC_i^\top$).

To solve this general class of convex programs, we reformulate problem \eqref{convProgGen} as 
\begin{align}\label{convProgGenRef}
&\minimize_{\bbeta\in\R^p,\vz_1,\ldots,\vz_n\in\R^{p+1}} ~~ \sum_{i=1}^n ~ \phi(\vz_i)~~~~\mbox{subject to:}~~~~\tilde\mC_i\bbeta - \vz_i=\tilde\va_i,~~i\in[n], 
\end{align}
where 
\begin{equation*}
 \phi(\vz) = \frac{1}{2}(|z_1|+\gamma\|\vz_2\|)^2, ~\mbox{for}~~\vz = \begin{bmatrix}z_1\\ \vz_2\end{bmatrix}\in \R^{p+1},  
~~\tilde\mC_i = \begin{bmatrix}\vb_i^\top \\ \mC_i^\top \end{bmatrix}, ~~\tilde\va_i = \begin{bmatrix} a_i\\ \boldsymbol{0}\end{bmatrix}.
\end{equation*}
Once the augmented Lagrangian is formed, a scaled ADMM scheme can be sketched to address \eqref{convProgGenRef} as follows (see \S3.1 \cite{boyd2011distributed}):\vspace{-.1cm} 
\begin{align}\notag 
    \vz_i^{k+1} &= \argmin_{\vz} ~ \phi(\vz) + \frac{\rho}{2}\!\left\|\vz - \tilde\mC_i\bbeta^k+\tilde\va_i+\vu_i^k \right\|^2\!\!, ~i\in[n],\\ \notag \bbeta^{k+1} &= \argmin_{\bbeta}~ \frac{\rho}{2}\sum_{i=1}^n \left\|\tilde\mC_i\bbeta - \vz_i^{k+1} - \tilde\va_i - \vu_i^k \right\|^2\!\!,\\ \notag \vu_i^{k+1} &= \vu_i^k - \tilde\mC_i\bbeta^{k+1} + \vz_i^{k+1} +\tilde\va_i, ~~i\in[n],\vspace{-.1cm} 
\end{align}
where $k$ represents the iteration index, $\rho>0$ is the penalty parameter, and $\vu_i$ represents the scaled dual vector corresponding to the $i$-th equality constraint in \eqref{convProgGenRef}.  
Assuming that $\sum_{i=1}^n\tilde\mC_i^\top\tilde\mC_i$ is full-rank, the proposed ADMM scheme reduces to the following chain of updates:\vspace{-.1cm} 
\begin{align}\notag 
    \vz_i^{k+1} &= \prox_{\rho^{-1}\phi(.)}\left( \tilde\mC_i\bbeta^k-\tilde\va_i-\vu_i^k\right), ~~i\in [n],\\ \notag \bbeta^{k+1} &= \big (\sum_{i=1}^n \tilde\mC_i^\top\tilde\mC_i\big )^{-1}\sum_{i=1}^n\tilde\mC_i^\top \left(\vz_i^{k+1} + \tilde\va_i + \vu_i^k \right),\\ \label{ADMMScheme} \vu_i^{k+1} &= \vu_i^k - \tilde\mC_i\bbeta^{k+1} + \vz_i^{k+1} +\tilde\va_i, ~~ i\in [n].\vspace{-.1cm} 
\end{align}
We would like to add few remarks about the proposed computational scheme. Regarding the $\bbeta$-update step, the inverse operator of $\sum_{i=1}^n\tilde\mC_i^\top\tilde\mC_i$, only needs to be calculated \emph{once} at the beginning of the algorithm (and clearly, it is numerically more stable to be performed through the calculation of the Cholesky decomposition instead of an explicit inverse matrix). Moreover, in many real-world problems, the number of missing patterns is much less than $n$, and
the number of factorizations to obtain the matrices $\mC_i$ in the aforementioned scheme turns out to be much less than the number of samples. It is also worth mentioning that the proposed scheme can be easily modified to a stochastic or mini-batch version for large-scale problems, by replacing the $\bbeta$-update with\vspace{-.1cm} 
\[ \bbeta^{k+1} = \big (\sum_{i\in B_k} \tilde\mC_i^\top\tilde\mC_i\big )^{-1}\sum_{i\in B_k}\tilde\mC_i^\top \left(\vz_i^{k+1} + \tilde\va_i + \vu_i^k \right),\vspace{-.2cm} 
\]
where $B_k\subset[n]$ is a batch selected at the $k$-th iteration. 

Furthermore, with regards to the $\vz$-update step, although an ADMM scheme can operate with a wide range of penalty parameters $\rho$, simple varying penalty schemes (specifically the scheme in \S3.4.1 \cite{boyd2011distributed}) can help with automating the selection of $\rho$, and speeding up the convergence in practice. Finally, in addition to the natural distributability of the $\vz$-update step among parallel computing units, it can be further simplified by acquiring a closed-form expression for the proximal operator of $\phi$. It is noteworthy that acquiring such closed-form expression requires a careful step-by-step analysis, which is performed in the proof of Theorem \ref{mainProxTh} below. However, the end result stated below, is computationally concise and easily accessible to use. 
\begin{theorem}\label{mainProxTh}
Given $\lambda\geq 0$ and $\gamma\geq 0$, consider $(z_1^*,\vz_2^*)$ to be the solution to the proximal operator \vspace{-.1cm} 
\begin{equation*}\label{mainProxThEq}
    \minimize_{z_1\in\R, \vz_2\in\R^p}  ~\frac{\lambda}{2}(|z_1|+\gamma\|\vz_2\|)^2 + \frac{1}{2 }(z_1-z_1')^2 + \frac{1}{2}\left\|\vz_2 - \vz_2' \right\|^2.\vspace{-.1cm} 
\end{equation*}
Then $z_1^*$ and $\vz_2^*$ are characterized as:
\[\left\{\begin{array}{lll} \mbox{if}~~ \lambda\gamma\|\vz_2'\|\geq (1+\lambda\gamma^2)|z_1'|: & z_1^* = 0, & \vz_2^* = \frac{1}{1+\lambda\gamma^2}\vz_2' \\
\mbox{if}~~ \frac{\lambda^2\gamma^2 |z_1'|}{\lambda+1}\leq \lambda\gamma \|\vz_2'\| < (1+\lambda\gamma^2)|z_1'|: & z_1^* = \frac{(1+\lambda\gamma^2)z_1' - \lambda\gamma \sign(z_1')\|\vz_2'\|}{1+\lambda+\lambda\gamma^2}, & \vz_2^* = \frac{(\lambda+1)\|\vz_2'\|-\lambda\gamma |z_1'|}{(1+\lambda+\lambda\gamma^2)\|\vz_2'\|} \vz_2'\\
\mbox{if}~~ \lambda\gamma\|\vz_2'\|<  \frac{\lambda^2\gamma^2 |z_1'|}{\lambda+1} : & z_1^* = \frac{z_1'}{1+\lambda}, & \vz_2^* = \boldsymbol{0}
\end{array} \right.\!\!\!\!.
\]
\end{theorem}
\begin{proof}
See the Appendix section \ref{th1proof}.  
\end{proof}
One immediately observes that the calculation of the proximal operator ultimately amounts to simple operations (e.g., calculation of the $\ell_2$ norm of the input vector), which can be done in linear time. 
\section{Notes on the Behavior of the Robust Risk}\label{robust}
Inspired by \eqref{eq:defL} and \eqref{eq:minL}, in this section we present some discussions about asymptotic behavior of the RIGID loss. For this purpose, we still consider the regression model in Section \ref{PS}, while making the following assumptions:
\begin{itemize}
    \item[A.1] The noise model is $\epsilon\!\sim\! \mathcal{N}(0,\sigma^2)$. 
    \item[A.2] The samples $\vx\in\R^p$ are centered and follow the multivariate normal distribution  $\vx\sim\mathcal{N}(\boldsymbol{0},\bSigma)$, where $\bSigma\succ \boldsymbol{0}$. 
    \item[A.3] The missing entries of $\vx$, denoted by $\eM$, are independent of $\vx$, and can follow one of the $M$ preset patterns $\{\eM_1,\ldots,\eM_M\}$, with probability $\pi_j>0$, where $\eM_j\subseteq [p]$, and $\sum_{j=1}^M\pi_j=1$.
\end{itemize}
In assumption A.2, $\vx$ is considered centered to avoid unnecessary complex formulations, although the results of this section can be generalized to the non-centered samples as well. Moreover, in addition to the independence from $\vx$, the assumption about missing entries in A.3 is posed quite generally. For example, if the missingness of each sample is completely at random, we have $M=2^p$ and the set of preset patterns is the power set of $[p]$. 

Under these assumptions, we are interested in understanding the behavior of the robust risk defined as
\begin{equation} \label{RobustRisk}
     \mathcal{L}(\bbeta;\gamma)\triangleq ~\EE\! \max_{\vx_\eM\in\mathcal{R}_\eM(\bbeta;\gamma)} \frac{1}{2}\left (y - \!\vx_{\!\eM}^\top \bbeta_{\!\eM} - \vx_{\!\eM^c}^\top \bbeta_{\!\eM^c} \right)^2\!\!,
     \vspace{-.15cm} 
\end{equation}
where \vspace{-.1cm} 
\begin{equation*}
\mathcal{R}_\eM(\bbeta;\gamma ) = \left\{\vz:~\! \left |\vz^\top \bbeta_{\eM} - \bar\bmu_\eM^\top \bbeta_{\mathcal{M}} \right | \leq \gamma \left\| \bbeta_{\mathcal{M}}\right\|_{\bar \bSigma_\eM} \right\},\vspace{-.1cm} 
\end{equation*} 
and \vspace{-.1cm} 
\begin{equation*}
\bar \bmu_\eM=\bSigma_{\mathcal{M} \eM^c} \bSigma_{\eM^c \eM^c}^{-1} \vx_{\eM^c}, ~~~ \bar \bSigma_\eM = \bSigma \slash \bSigma_{\eM^c\eM^c}.\vspace{-.1cm} 
\end{equation*}
One could immediately see the connection between the robust risk \eqref{RobustRisk} and the RIGID formulation in \eqref{eq:defL} and \eqref{eq:minL}, when the number of samples $n$ is very large. In fact, using a similar line of argument as Proposition \ref{prop:Li}, it is straightforward to show that
\[\mathcal{L}(\bbeta;\gamma) = ~\EE \left( \left |y - \vx_{\eM^c}^\top \bbeta_{\eM^c} - \bar\bmu_\eM^\top \bbeta_{\mathcal{M}} \right|+ \gamma \left\| \bbeta_{\mathcal{M}}\right\|_{\bar \bSigma_\eM} \right)^2.
\]
We would like to note that in this equation (also in \eqref{RobustRisk}) the expectation is taken with respect to all the three sources of model randomness, $\vx, ~\eM$ and $\epsilon$. Before we proceed with deriving a closed-form expression for $\mathcal{L}(\bbeta;\gamma)$, we introduce a notation that helps with our presentation. 

For a missing index set $\eM\subseteq [p]$, consider $\mS_{\eM}\in\R^{|\eM|\times p}$ to be the selection matrix obtained by picking a subset of the rows of the identity matrix, where $\bbeta_\eM = \mS_\eM\bbeta$. Correspondingly we define 
 \begin{equation}\bar \bSigma_\eM^E = \mS_\eM^\top \bar \bSigma_\eM \mS_\eM.\label{eqSigmaE}
\end{equation}
In simple words, unlike $\bar \bSigma_\eM$ which is of size  $|\eM|\times|\eM|$, $\bar \bSigma_\eM^E$ is a $p\times p$ zero-padded matrix that has $\bar \bSigma_\eM$ in the submatrix indexed by $\eM\times \eM$, and contains zeros elsewhere. We are now ready to present a closed-form expression for $\mathcal{L}(\bbeta;\gamma)$. 
\begin{theorem}\label{th:LclosedForm}
Assume that A.1-A.3 hold, and $\mathcal{L}(\bbeta;\gamma)$ follows the formulation in \eqref{RobustRisk}. Then 
\begin{align}\notag 
 \mathcal{L}(\bbeta;\gamma) =   \sigma^2\! &+ \!\sum_{j=1}^M \!\pi_j\!\left( \left\|\bbeta-\bbeta_0\right\|^2_{\bSigma - {\bar \bSigma_{\eM_j}^E}} \!\!+\! \left\|\bbeta_0\right\|^2_{ {\bar \bSigma_{\eM_j}^E}} \!\!+ \!\gamma^2 \left\|\bbeta \right\|^2_{\bar \bSigma_{\eM_j}^E} \!\right)
 \\& \!+\!    \gamma\sqrt{\!\frac{8}{\pi}}\!\sum_{j=1}^M \!\pi_j \!\left\|\bbeta \right\|_{\bar \bSigma_{\eM_j}^E} \!\! \left( \sigma^2 \!+\! \left\|\bbeta-\bbeta_0\right\|^2_{\bSigma - {\bar \bSigma_{\eM_j}^E}} \! \!\!+\! \left\|\bbeta_0\right\|^2_{ {\bar \bSigma_{\eM_j}^E}}\! \right)^{\frac{1}{2}}\!\!\!\!.\label{th:eqLBeta}
\end{align}\vspace{-.4cm}
\end{theorem}
\begin{proof}
See \S~\ref{th2Prf} of the Supplementary Material. 
\end{proof}
A first question about \eqref{th:eqLBeta} is the positivity of the matrices $\bSigma - {\bar \bSigma_{\eM_j}^E}$ and ${\bar \bSigma_{\eM_j}^E}$, which appear as weighting matrices in the $\ell_2$ norm terms. Based on a well-established property of the Schur complement, given $\bSigma\succ \boldsymbol{0}$, for any non-empty pattern $\eM$, $\bar\bSigma_\eM$ is also positive definite. Furthermore, as detailed in the proof of Theorem \ref{th:robustRiskSparse} below, it is guaranteed that $\bSigma - {\bar \bSigma_{\eM}^E}$ and ${\bar \bSigma_{\eM}^E}$ are both positive semi-definite.

Further assessing \eqref{th:eqLBeta}, one observes that aside from $\sigma^2$, $\bbeta_0$ and $\gamma$, other factors such as the structure of the covariance matrix, $\bSigma$, the structure of the missing patterns, and their probability mass values, $\pi_j$, contribute to the characteristics of the robust risk. Because of the diversity of the controlling parameters, presenting general results about the behavior of $ \mathcal{L}(\bbeta;\gamma)$, without making strong assumptions about some of these parameters is challenging. It is however possible to characterize some interesting aspects of the solutions, as discussed in the sequel.

One of the main characteristics of RIGID is promoting sparse solutions. The close connection between model sparsity and robustness has been observed and investigated in the past \cite{guo2018sparse, NIPS2017_3fab5890, aghasi2020fast, gopalakrishnan2018combating}, and RIGID supports a similar idea. A common observation is increasing $\gamma$ in RIGID promotes sparser solutions, and specifically when the missingness is affecting all the feature components, operating beyond a finite value of $\gamma$ returns a zero solution. This is also a property observed in LASSO \cite{tibshirani1996regression}.   

\begin{theorem}\label{th:robustRiskSparse}
Assume that A.1-A.3 hold, and for $\gamma\geq 0$, $\bbeta^\gamma$ is a minimizer of $\mathcal{L}(\bbeta;\gamma)$ defined in \eqref{RobustRisk}: 
    \begin{itemize}
    \item[(a)] For $\gamma=0$, $\bbeta^\gamma$ is unique if (and only if) 
$\sum_{j=1}^M\pi_j\left(\bSigma-\bar\bSigma_{\eM_j}^E \right)\succ \boldsymbol{0}$. 

    \item[(b)] For any $\gamma>0$, $\mathcal{L}(\bbeta;\gamma)$ is strictly convex, and $\bbeta^\gamma$ is unique. 
       
    \item[(c)]  If the condition in (a) holds, then as $\gamma\to 0$, $\bbeta^\gamma$ converges  to $\bbeta_0$.\vspace{-.05cm}

    \item[(d)] Define the set $J = \{j\in[M]: \sigma + \|\bbeta_{0\eM_j}\|>0\}$, and let $\underline{\lambda}_j$ denote the smallest eigenvalue of $\bar\bSigma_{\eM_j}$ (which is guaranteed to be strictly positive when $\bSigma\succ\boldsymbol{0}$). If $\bigcup_{j\in J}\eM_j= [p]$, then there exists a finite-valued $\gamma_0$, such that $\forall \gamma\geq \gamma_0,~\bbeta^\gamma = \boldsymbol{0}$. One choice for $\gamma_0$ (not necessarily the smallest) is 
    \begin{equation}
        \gamma_0 =  \left\| \frac{\sqrt{\pi}}{\kappa_{\min}} \sum_{j=1}^M\pi_j\left(\bSigma - \bar\bSigma_{\eM_j}^E \right) \bbeta_0 \right\|,
    \end{equation}
    where $\kappa_{\min} = \min_{j\in J}~~\pi_j \underline{\lambda}_j \left( \sigma +  \underline{\lambda}_j\left\|\bbeta_{0\eM_j}\right\|\! \right)$.
\end{itemize}
\end{theorem}
\begin{proof}
See the Appendix section \ref{th3proof}.
\end{proof}
Showing the strict convexity of $\mathcal{L}(\bbeta;\gamma)$ requires a careful analysis as detailed in the proof.  With regards to part (d) of the Theorem, in real-world problems that $\sigma>0$, the index set $J$ is simply $[p]$. The theorem basically states that when every feature in the dataset appears as an element of at least one missing pattern, increasing $\gamma$ beyond a finite value can produce a zero solution. This behavior can be ceased if the union of the missing patterns does not cover all the $p$ features. The proposition below proposes a problem setup in which the missing patterns do not cover all the features, and RIGID does not return an all-zero solution no matter how large $\gamma$ is. 
\begin{proposition}\label{prop:SinglePattern}
	Assume that A.1 and A.2 hold, and A.3 holds for $M=2$, where $\pi_0\in(0,1]$ portion of the data is complete and the remaining $1-\pi_0$ follows a single missing pattern $\mathcal{M}$. Denote the minimizer of $\mathcal{L}(\bbeta;\gamma)$, by $\bbeta^\gamma$, then:
    \begin{itemize}
    \item[(a)] For every $\gamma\geq 0$, $\bbeta^\gamma$ is unique. \vspace{-.05cm} 
    \item[(b)] The unique solution $\bbeta^\gamma$ obeys 
        $\bbeta^\gamma_{\eM^c} = {\bbeta_0}_{\eM^c} + \bSigma_{\eM^c\eM^c}^{-1}\bSigma_{\eM^c \eM}{\bbeta_0}_{\mathcal{M}}$ and  $\bbeta^\gamma_{\mathcal{M}}=\boldsymbol{0} 
        $, if 
        \[\gamma\geq \sqrt{\frac{\pi}{2}} \frac{\pi_0 \left\|\bbeta_{0\eM}\right\|_{{\bSigma\slash \bSigma}_{\eM^c\eM^c}} }{  (1-\pi_0)   \left( \sigma^2 + \left\|\bbeta_{0\eM}\right\|^2_{{\bSigma\slash \bSigma}_{\eM^c\eM^c}}\! \right)^{\frac{1}{2}}}. 
        \] 
\end{itemize}
\end{proposition}
\begin{proof}
See \S~\ref{ProofOfExampleAsym} of the Supplementary Material. 
\end{proof}
We can see that in this case increasing $\gamma$ can still cause the components of the solution that appear in the missing pattern to vanish to zero, however, the remaining components are unaffected by such shrinkage. We also observe that as the portion of missing data increases (i.e., $\pi_0$ decreases), the vanishing of the coefficients corresponding to the missing features happens at a smaller $\gamma$ value. In a sense, RIGID has a stronger tendency towards dropping the missing patterns (or features) with larger missing rate.

\section{Estimation of the Mean and Covariance}\label{sec:EstSufSt}
So far, in our formulations we assumed that $\vx_i\sim\mathcal{N}(\bmu,\bSigma)$, where both $\bmu$ and $\bSigma$ are known. However, in many problems of interest, these parameters are unknown and need to be estimated from the data with missing entries. This section discusses the process of estimating these quantities, and provides concentration bounds on how they relate to the actual parameters.

A main component of sample mean and sample covariance in full data regimes is normalization the number of samples, $n$. The basic idea in the missing data regime is a normalization by the number of available components. For this purpose, we define
\begin{equation*}
p_{jk} = \frac{1}{n}\sum_{i=1}^n 1_{j\in\Ai}1_{k\in\Ai}, ~~~k, j\in[p]. 
\end{equation*}
Indeed, $p_{jk}$ is the ratio of samples for which both feature components indexed by $j$ and $k$ are available. In the complete data regime, we trivially would have $p_{jk}=1$.

In the missing regime, we can estimate $\bmu$ by $\hat\bmu$, whose $j$-th element is calculated as 
\begin{equation*}
\hat\mu_j = \frac{\sum_{i=1}^n x_{i,j}1_{j\in\Ai}}{\sum_{i=1}^n 1_{j\in\Ai}} = \frac{\sum_{i=1}^n x_{i,j}1_{j\in\Ai}}{np_{jj}}.  
\end{equation*}
Here $x_{i,j}$ indicates the $j$-th component of $\vx_i$. Basically, the $j$-th element of $\hat\bmu$ is calculated by summing over the $j$-th element of the samples which have it available, and normalizing the sum by the number of samples which have that entry available. Trivially, the proposed formulation reduces to the standard sample mean formulation in the complete data regime. 

In a similar fashion, the covariance calculation can also be restricted to the available samples. More specifically, the true covariance matrix $\bSigma$ can be estimated by the matrix $\hat\bSigma$, whose elements are calculated as  
\begin{align*}
\hat\Sigma_{jk} &= \frac{\sum_{i=1}^n (x_{i,j}-\hat\mu_j)(x_{i,k}-\hat\mu_k)1_{j\in\Ai}1_{k\in\Ai}}{\sum_{i=1}^n 1_{j\in\Ai}1_{k\in\Ai}} 
= \frac{\sum_{i=1}^n (x_{i,j}-\hat\mu_j)(x_{i,k}-\hat\mu_k)1_{j\in\Ai}1_{k\in\Ai}}{np_{jk}} .\vspace{-.1cm} 
\end{align*}
Again the formulation simplifies to the standard sample covariance formulation in the complete data regime.  It is straightforward to show that $\hat\bmu$ is an unbiased estimate of $\bmu$. While the covariance estimate exhibits some level of bias as detailed below, it is negligible to an extent that works such as \cite{cai2016minimax} refer to it as \emph{nearly unbiased}. The following result sheds light on this.
\begin{proposition} \label{propbias}For the estimates $\hat\bmu$ and $\hat\bSigma$ proposed above, one has $\EE\hat\bmu = \bmu$, and  
\begin{equation}\label{eq6}\EE\hat \Sigma_{jk} = \Sigma_{jk}\left( 1 - \frac{p_{jj}+p_{kk}-p_{jk}}{np_{jj}p_{kk}}\right). 
\end{equation}
\end{proposition}
\begin{proof}
See the Appendix section \ref{prop3proof}.
\end{proof}
\noindent It is immediately observable that in the complete data regime, \eqref{eq6} reduces to the well-known result $\EE\hat\bSigma=(1-n^{-1})\bSigma$. 

Unlike the sample covariance matrix which is always positive semi-definite (PSD), $\hat\bSigma$ is symmetric but not guaranteed to be PSD. To ensure the positivity of the estimate, we suggest using a projected matrix $\hat\bSigma^+$ calculated via 
\begin{equation}\label{eq7}
\hat\bSigma^+ = \argmin_{\mS\ \!\succeq\ \!\boldsymbol{0}} ~\left \|\mS-\hat\bSigma\right \|. 
\end{equation}
In fact, to avoid issues related to calculating the Schur complement of designated blocks in $\hat\bSigma^+$, we can replace the $\mS\ \!\succeq\ \!\boldsymbol{0}$ constraint in \eqref{eq7} with $\mS\ \!\succeq\ \!\theta\mI$, where $\theta$ is a small positive number. Since a high condition number of the covariance matrix can increase the model variance, $\theta$ can be taken large enough to the extent that the  condition number of $\hat\bSigma^+$ stays below a certain cap. This way the estimated covariance and the Schur complement of any of its diagonal blocks are guaranteed to be well-conditioned and positive definite. Interestingly, using the well-established result of \cite{halmos1972positive} (or see Theorem 3.2 of \cite{higham1988matrix}) we can precisely characterize the solution to \eqref{eq7}:
\begin{proposition}\label{propproj}
	Denoting the smallest eigenvalue of $\hat \bSigma$ by $\lambda_{\min}(\hat\bSigma)$, the convex program \eqref{eq7} admits the closed-form solution  $\hat\bSigma^+ = \hat\bSigma - \min(0,\lambda_{\min}(\hat\bSigma))\mI$. 
\end{proposition}	
\begin{proof}
See the Appendix section \ref{prop4proof}.
\end{proof}
A valid question is how the distance between the proposed estimates and the actual parameters shrinks with the dimension, the number of samples and the missing data ratios $p_{jk}$. As detailed in the theorem below, in the missing data regime, an effective replacement for $n$ is $np_{\min}$, where $p_{\min}$ is the minimum value of $p_{jk}$.

\begin{theorem}\label{thConc}
Consider the proposed mean and covariance estimates $\hat\bmu$ and $\hat\bSigma^+$, and the missing patterns to be independent of the data $\{\vx_i\}_{i\in [n]}$. Assume $\min_{j,k\in[n]}p_{jk}:=p_{\min}>0$, 
and fix $\nu>0$. Then,\vspace{-.2cm}
\begin{itemize}
\item[(a)] With probability exceeding $1 - 2\exp\left(-c\nu\right)$, where $c>0$ is a universal constant: 
$
\|\hat\bmu - \bmu\|^2 \leq \left(\trace(\bSigma) + \max\left({\nu}\|\bSigma\|, \sqrt{{\nu}}\|\bSigma\|_F \right)\right)/(np_{\min}).
$
\item[(b)] With probability exceeding $1-2\exp(-\nu)$: 
\begin{align*}
\left \|\hat\bSigma^+ - \bSigma \right \| &\leq \left(  \zeta(p,np_{\min},\nu)\|\bSigma\| - \lambda_{\min}(\bSigma)\right) ^+\\&~~~~~ + \zeta(p,np_{\min},\nu)\|\bSigma\|,\vspace{-.1cm} 
\end{align*}
where $u^+ = \max(u,0)$, and for positive universal constants $c_1, c_2$ and $c_3$, $ \zeta(p,np_{\min},\nu)\! =\! \max(\!(\frac{c_1+c_2p+\nu}{c_3np_{\min}})^{\frac{1}{2}}\!, \frac{c_1+c_2p+\nu}{c_3np_{\min}}\!). $
\end{itemize}
\end{theorem}
\begin{proof}
See the Appendix section \ref{th4proof}.
\end{proof}
The proposed procedure to estimate the covariance matrix is of practical value in $p<n$ regimes. For high-dimensional covariance estimation, where $p$ is large compared to the sample size $n$, and the data are incomplete, the reader is referred to some results and discussion presented in \S~4 of the Supplementary Material.

\section{Experiments and Discussions}\label{numerical}
In this section we present some numerical experiments related to the performance of RIGID in general, and in comparison with other techniques. Our experiments are performed on various types of data, including synthetic, semi-synthetic, and real data. Some of the most well-known techniques such as mean imputation, K-nearest neighbors imputation (KNN), imputation with conditional mean, multiple imputation by chain equations (MICE), and Amelia are used to impute the data, followed by a linear regression step. It is worth mentioning that while MICE is a rather general imputation framework, Amelia is specifically designed for multivariate normal data \cite{BuGr11, honaker2011amelia}. Among the methods which directly perform linear regression on incomplete data, the two most relevant and best performing ones turned out to be the sparse recovery in \cite{RosTsy09}, and the high dimensional regression with imputation work by Chandrasekher, Alaoui, and Montanari \cite{chandrasekher2020imputation}. We respectively refer to them as RT, and CAM (author abbreviated). To abide by the page limit, some details of the experiments, plus additional results are moved to \S~\ref{app:MoreExp} of the Supplementary Material.  

\begin{figure*}[htb!]
\centering \begin{overpic}[trim={0 -.25cm  0 0},clip,height=1.75in]{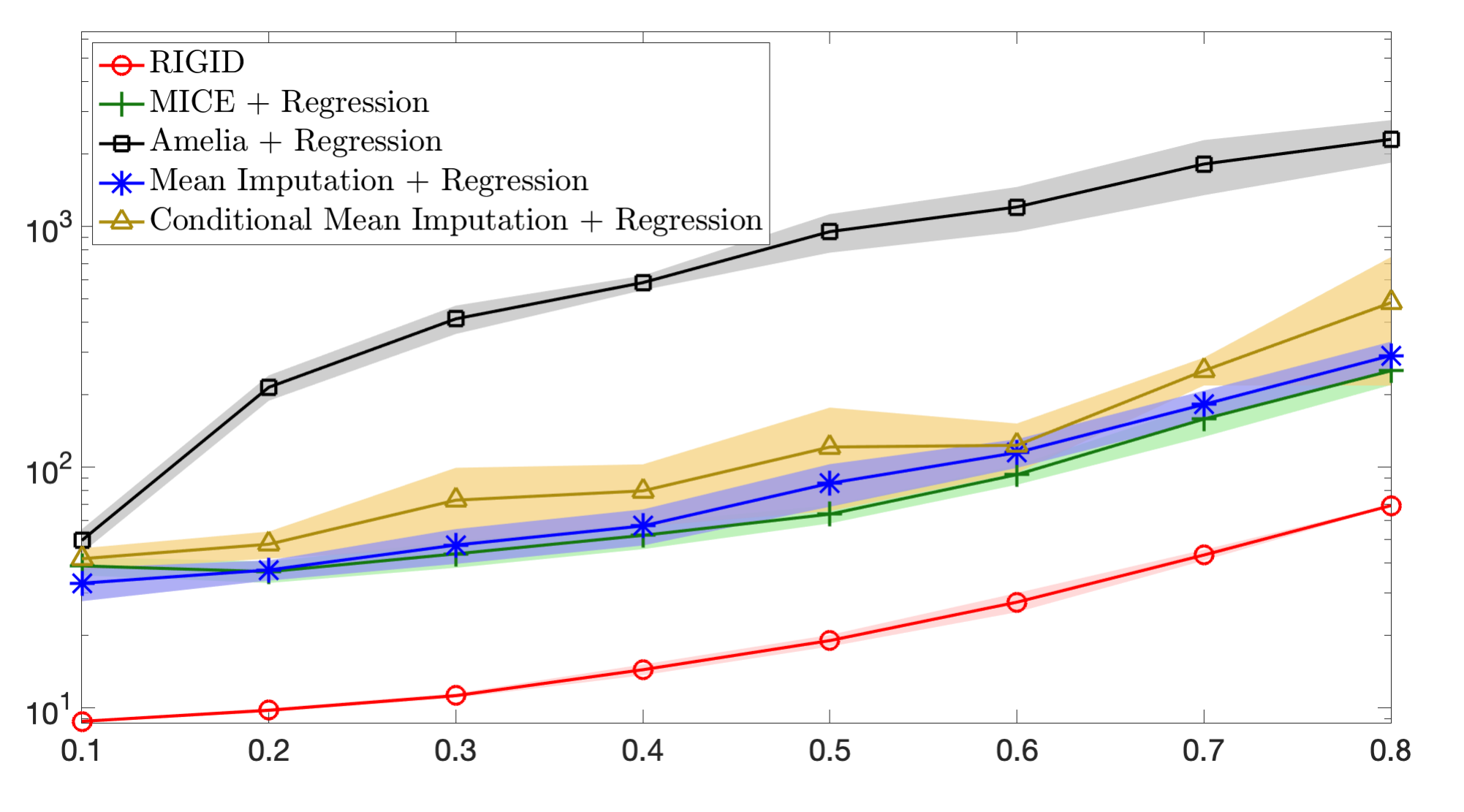}

\put (40,1) {\scalebox{.8}{Missing Ratio}} 
\put (99,-3) {\scalebox{.8}{(a)}} 

\put (-1,25) {\scalebox{.75}{\rotatebox{90}{RMSE}}} 
\end{overpic}\hspace{-.2cm}
\begin{overpic}[trim={0 -.25cm  0 0},clip,height=1.75in]{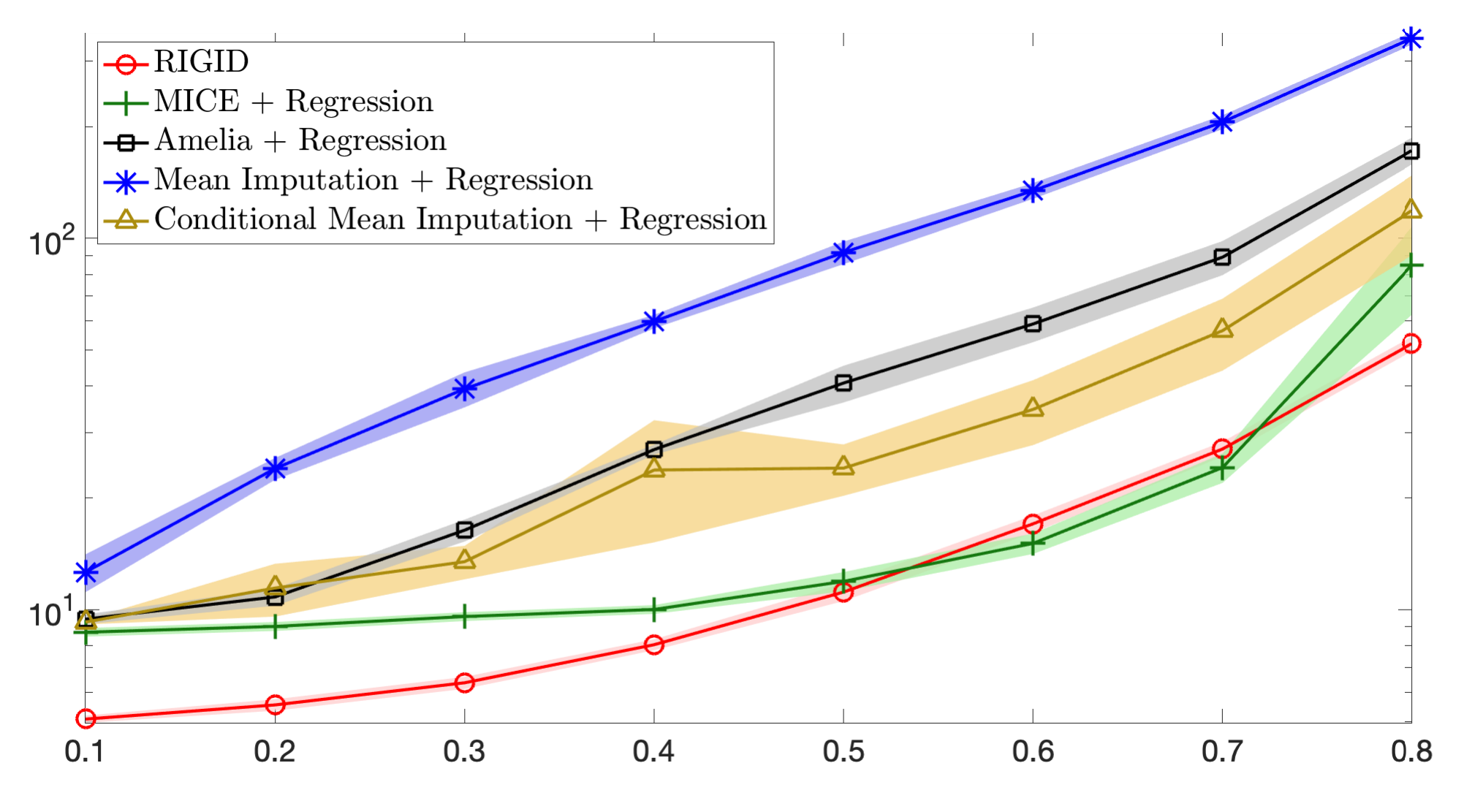}

\put (40,1) {\scalebox{.8}{Missing Ratio}}

\put (0,25) {\scalebox{.75}{\rotatebox{90}{RMSE}}} 
\end{overpic}\\[.3cm]
\begin{overpic}[trim={0 -.25cm  0 0},clip,height=1.75in]{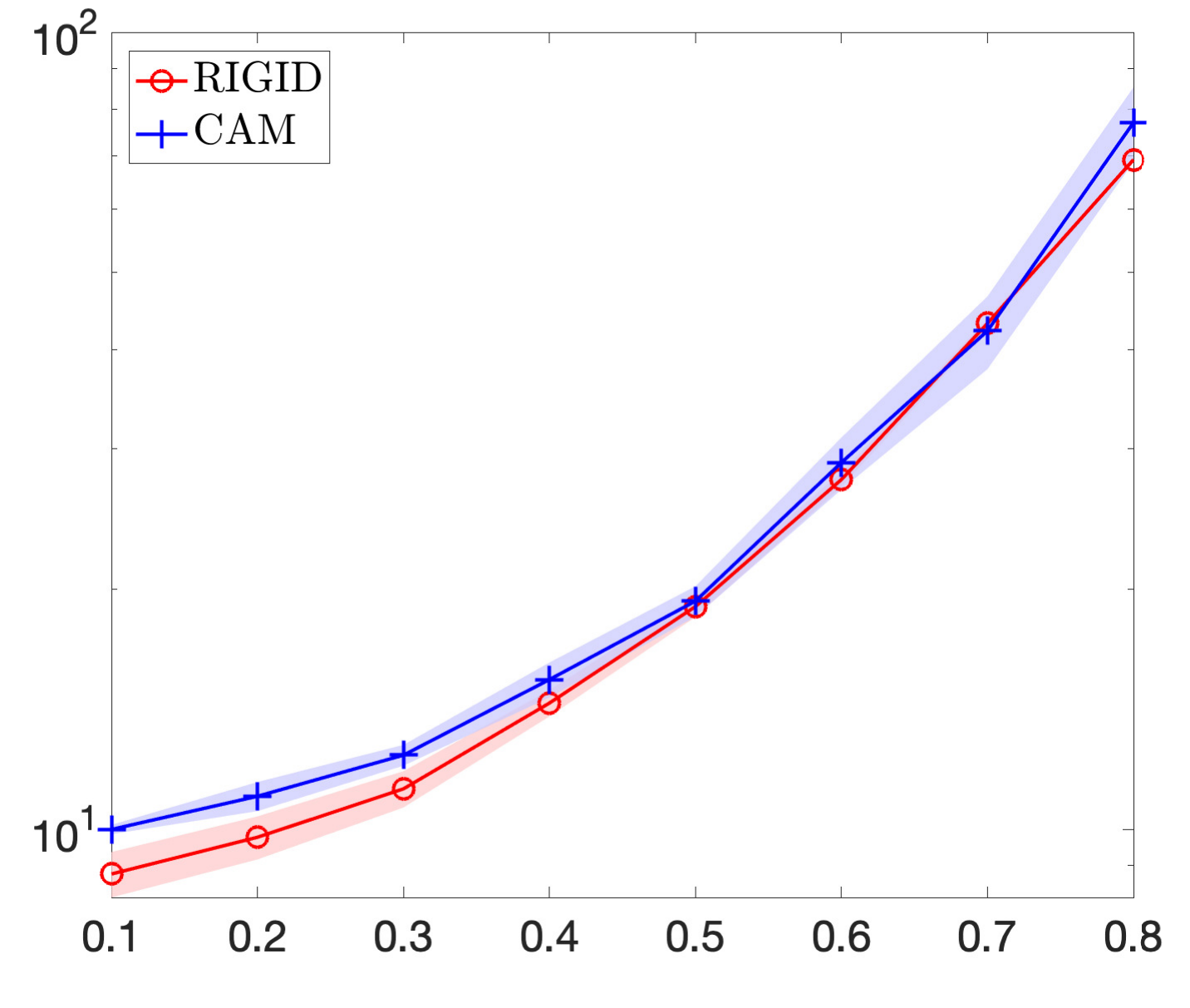}

\put (40,0.5) {\scalebox{.8}{Missing Ratio}}

\put (0,40) {\scalebox{.75}{\rotatebox{90}{RMSE}}} 
\end{overpic}
\vspace{.2cm}
\begin{overpic}[trim={0 -.25cm  0 0},clip,height=1.75in]{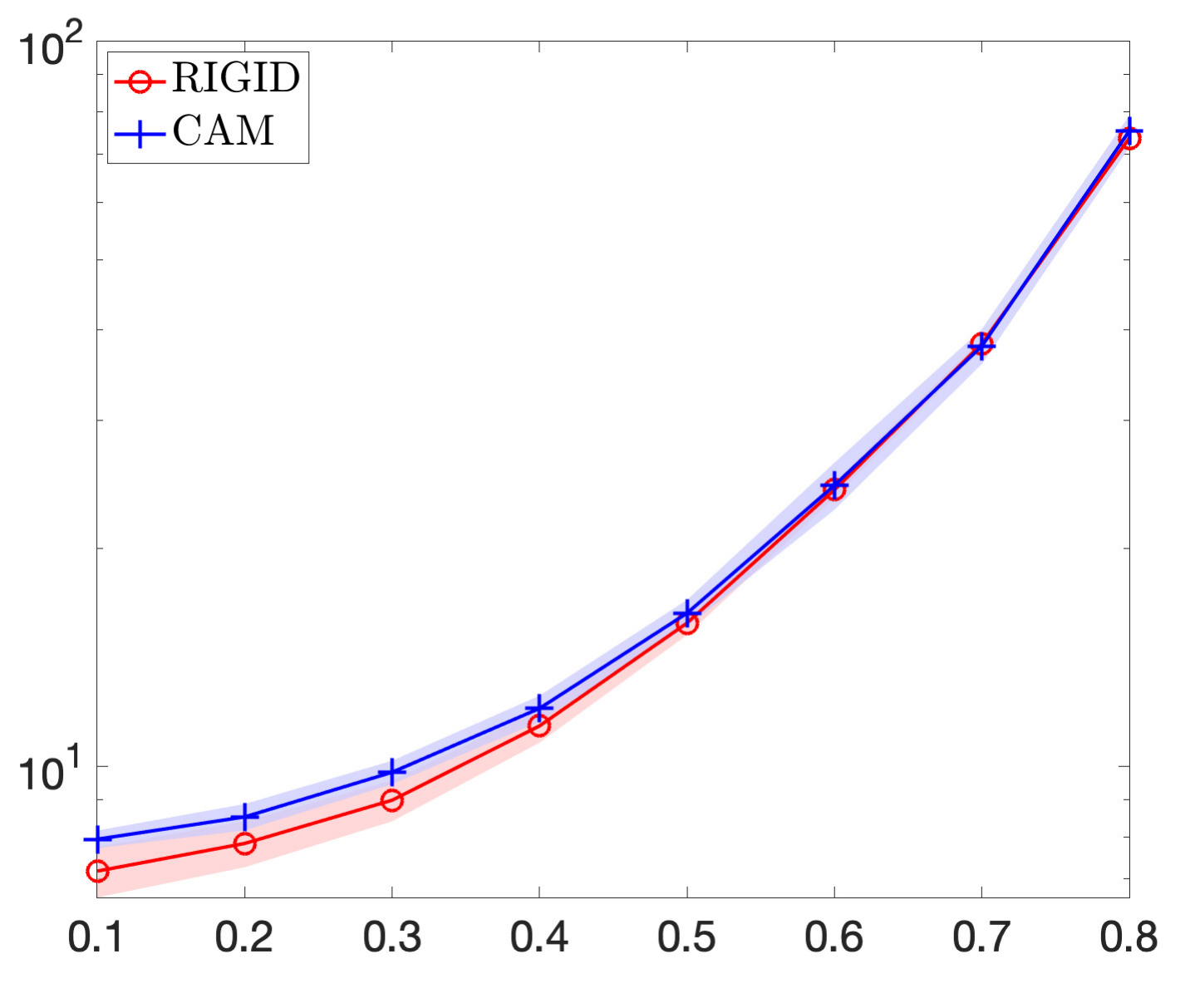}

\put (40,0.5) {\scalebox{.8}{Missing Ratio}} 
\put (50,-5.5) {\scalebox{.8}{(b)}} 

\put (0,40) {\scalebox{.75}{\rotatebox{90}{RMSE}}} 
\end{overpic}
\begin{overpic}[trim={0 -.25cm  0 0},clip,height=1.75in]{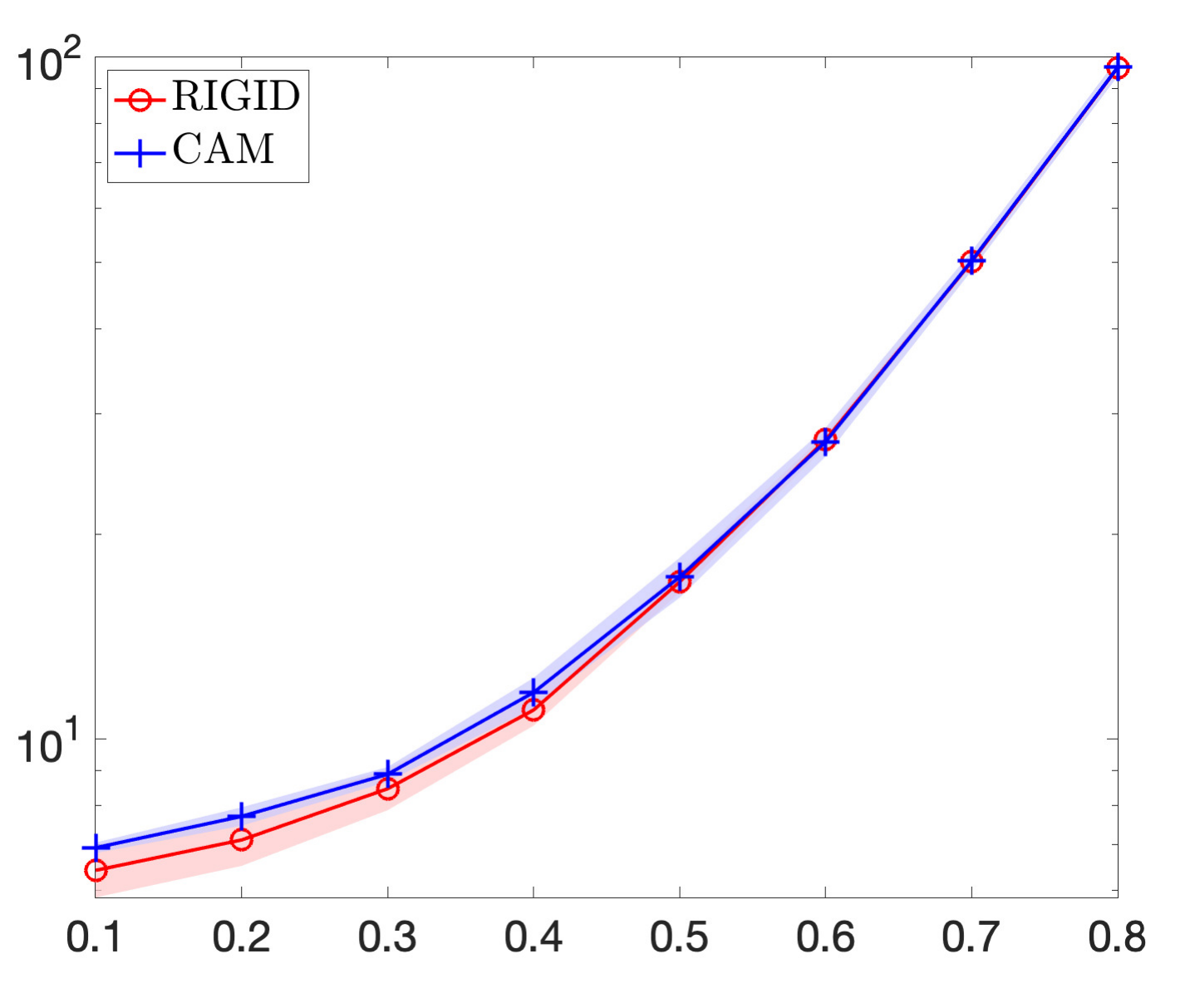}

\put (40,0.5) {\scalebox{.8}{Missing Ratio}}

\put (0,40) {\scalebox{.75}{\rotatebox{90}{RMSE}}} 

\end{overpic}\vspace{.2cm}

\centering \begin{overpic}[trim={0 -.25cm  0 .5cm},clip,height=2.1in]{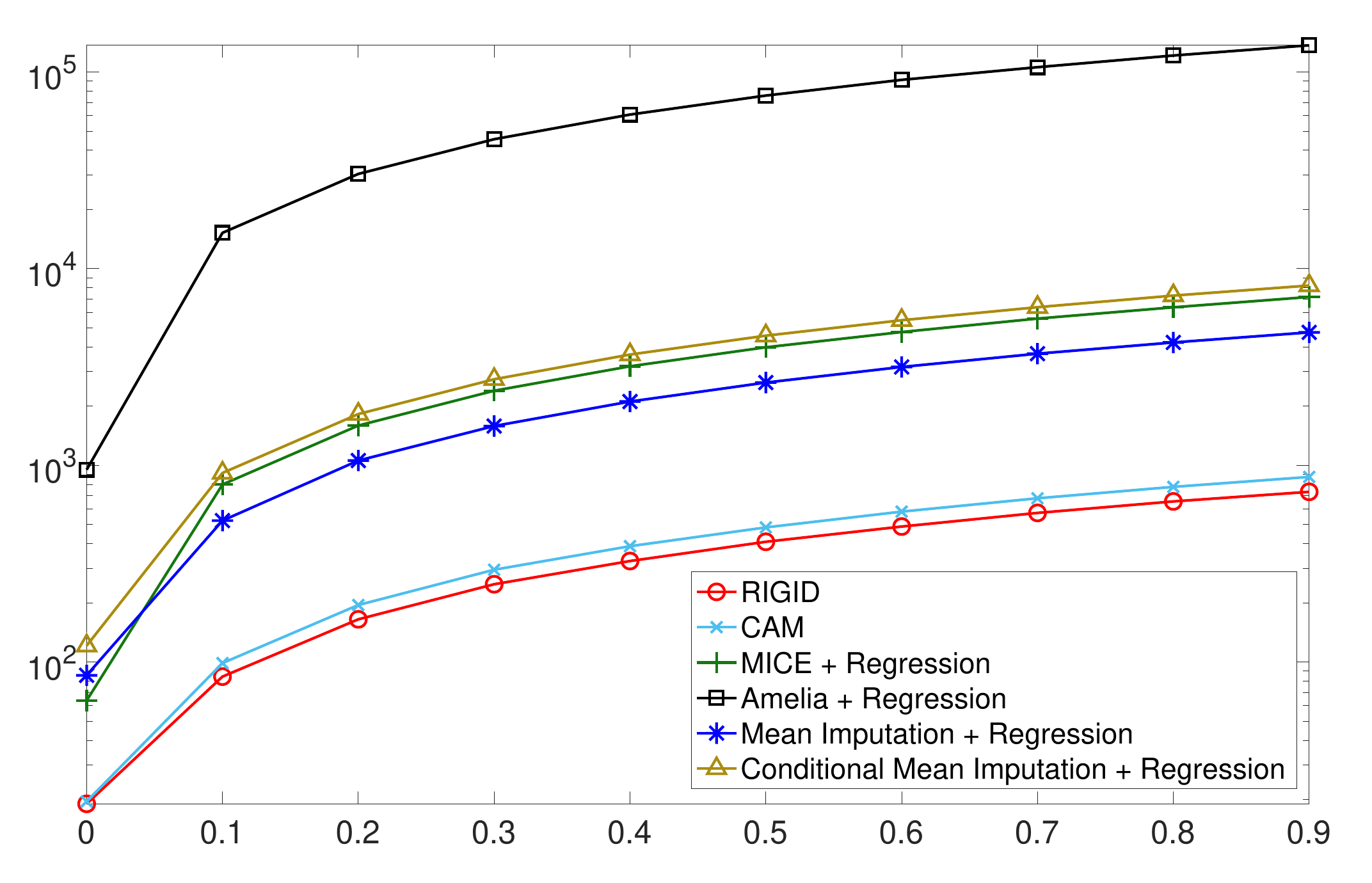}
\put (10,-1) {\scalebox{.8}{Ratio of Added Noise to Features ($\vx$): MCAR 50\%}} 
\put (-1,25) {\scalebox{.75}{\rotatebox{90}{RMSE}}} 
\put (50,-4.5) {\scalebox{.8}{(c)}} 
\end{overpic}\hspace{-.2cm}
\begin{overpic}[trim={0 -.25cm  0 .71cm},clip,height=2.1in]{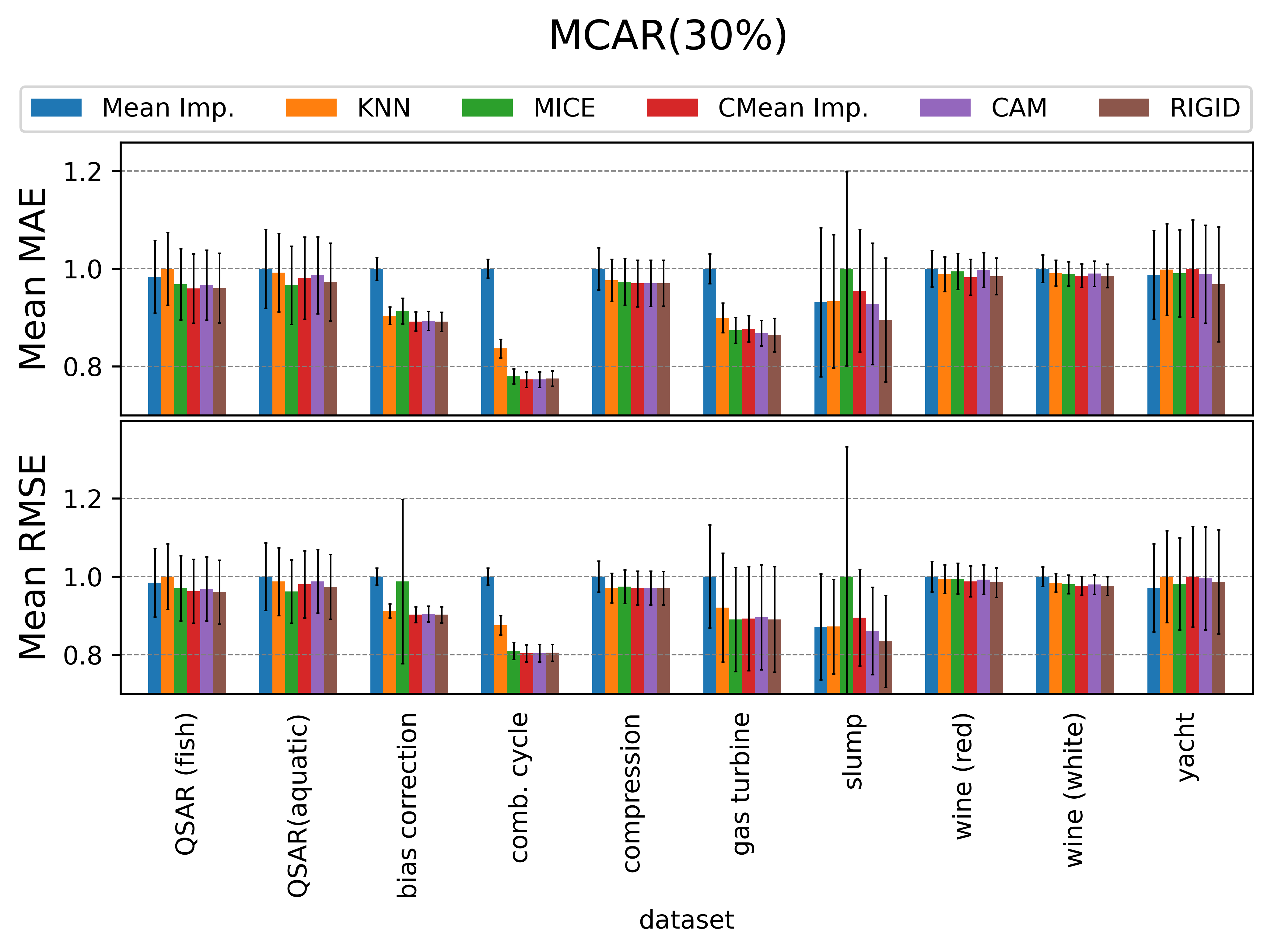}

\put (50,-4.5) {\scalebox{.8}{(d)}} 

\end{overpic}\\[.3cm]
 \caption{Comparing RIGID with other techniques. (a) RIGID vs imputation + regression techniques, left: $n=500$, right: $n=1000$. (b) RIGID vs CAM, left: $n=500$, middle: $n=1000$, right: $n=2000$. (c) Robustness to added noise to the features. (d) Comparison with real-world, non-Gaussian datasets. }\label{figSemi}\vspace{-.5cm}
\end{figure*}
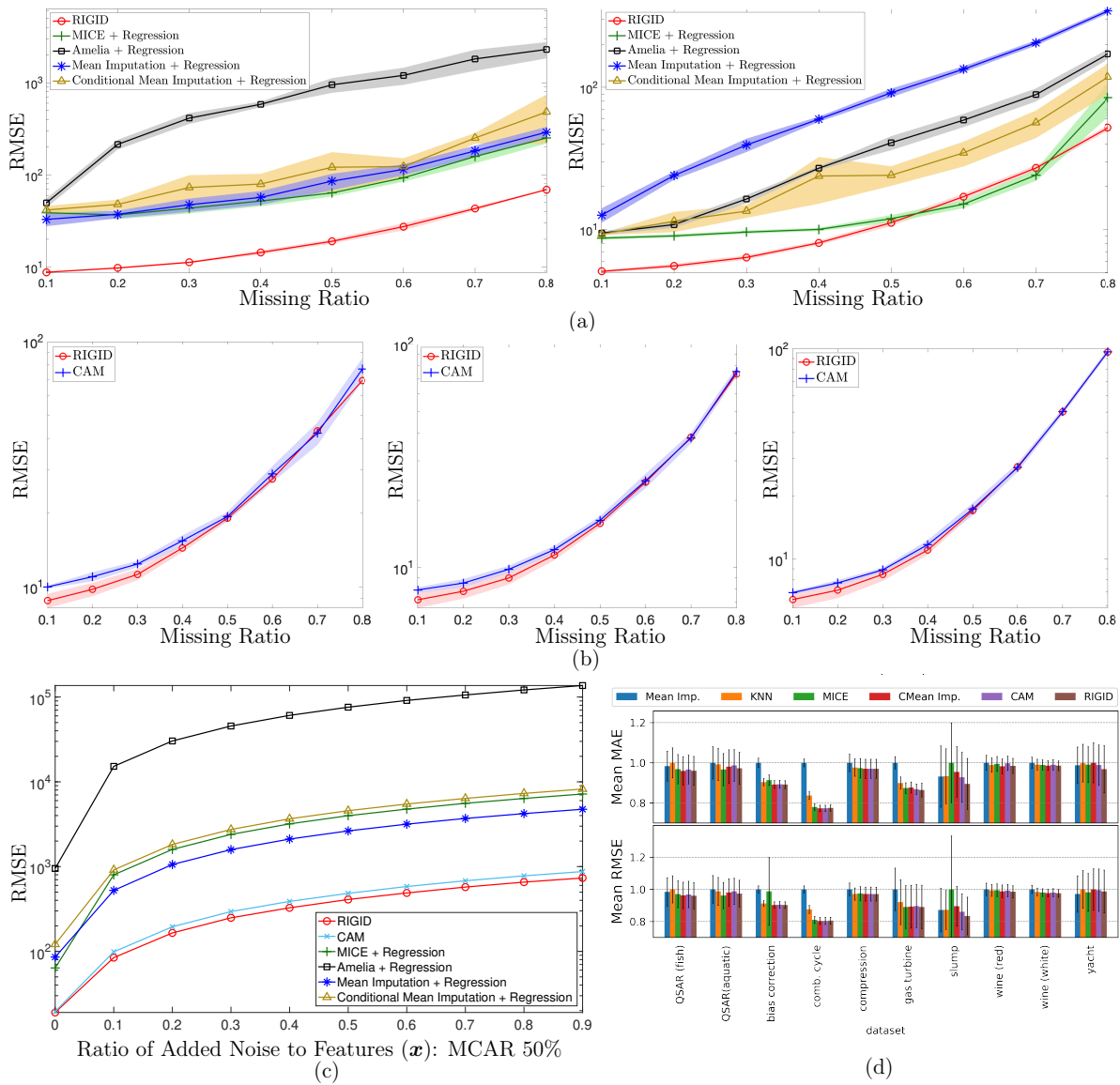

For our first set of experiments (semi-synthetic), we extracted the daily closing stock value of the companies which consistently appeared on the S\&P 500 list between 2012 and 2017. These data were used to create a covariance matrix and a mean vector for extensive multi-normal data generation. The details of this process along with the incorporation of noise and outliers are presented in the designated Supplementary section.
For these experiments we consistently have $p=447$, while $n$ varies between 500, 1000 and 2000. Panel (a) of Figure \ref{figSemi} presents the average root-mean-square error (RMSE) of the predictions for various missing ratios (that is, the total number of missing values divided by the total number of data points). Each experiment is executed several times, and the shaded region around each plot represents one-standard deviation confidence interval. We would like to note that MICE and Amelia become extremely slow in high dimensions (making the extensive experiments for $n=2000$ infeasible).

One can see the notable contrast between RIGID's performance, and impute + regress techniques, both in terms of accuracy and its variation. Panel (b) presents a comparison of RIGID and CAM for different values of $n$, where the gap is smaller, however still a better performance of RIGID is observed thanks to its customized formulation taking into account the feature dependency and data covariance matrix. For very large missing rates and large values of $n$, the gap between RIGID and CAM starts to shrink. Panel (c) compares the robustness of the models to uncertainty in the features. Thanks to its robust formulation, one can see that RIGID presents more robustness to feature uncertainty compared to the other competing techniques.  

In our second experiment, we compare the performance of our approach with RT \cite{RosTsy09} for solving problem \eqref{eq:minL} when $\bbeta_0$ is a sparse vector. The RT framework is specifically designed for recovering sparse solutions in incomplete data. The multi-normal data in this case is fully synthetic. In particular, we use two sets of observations, namely, $(n,p)=(500,100), (1000,500)$, and for each, we generate synthetic data by considering three levels of missing ratios, $0.25, 0.5, 0.75$, and four levels of sparsity for $\bbeta_0$, $s=0.8, 0.6, 0.4, 0.2$, where $s$ denotes the rate of non-zero elements in $\beta$ (larger $s$ indicates more zeros). After fine-tuning the parameter $\gamma$ for RIGID and another parameter for the sparse recovery algorithm, we run both algorithm $10$ times. The output solutions are then evaluated over a complete test data set of size $10000$. We then report the average and standard deviation of the RMSE in Table~\ref{RMSE_sparse}. For most cases when $s<0.8$, RIGID has a better performance than RT in terms of the average RMSEs, while it has slightly worse standard deviations. On the other hand, in large sparsity regimes where the optimal solution is sparser ($s=0.8$), the RT sparse recovery algorithm performs better as it has been specifically designed for sparse problems.
\begin{table}[t]
\caption{Comparison of RIGID and RT \cite{RosTsy09}. For each parameter setting, the ``mean'' and ``standard deviation'' cells show the average and deviation of RMSE over 10 runs of the algorithms.}
\vspace{0.1cm}
\centering
\label{RMSE_sparse}
\scriptsize
\resizebox{6cm}{!}{%
\begin{tabular}{|c|c|c|c|c|c|}
\hline
\multicolumn{6}{|c|}{\boldmath{$n=500, d=100$}}\\
\hline
\multicolumn{6}{|c|}{Missing ratio: $0.25$}\\
\hline
\multicolumn{2}{|c|}{Sparsity (s)}&$0.8$&$0.6$&$0.4$&$0.2$\\
\hline
\multirow{2}{*}{RIGID}& mean	&0.90	&1.83 & \bf{1.81}	&\bf{2.06}		\\
&std	&0.20	&0.45 &0.41	&0.42	\\
\hline
\multirow{2}{*}{RT}& mean	&\bf{0.62}	&\bf{1.71} &1.86	&2.23		\\
&std	&0.10	&0.31	&0.33 & 0.37	\\
\hline
\multicolumn{6}{|c|}{Missing ratio: $0.5$}\\
\hline
\multirow{2}{*}{RIGID}& mean	&1.66	&2.35	&\bf{2.49} & \bf{2.74}		\\
&std	&0.28	&0.47	&0.46 & 0.56	\\
\hline
\multirow{2}{*}{RT}& mean	&\bf{1.39}	&\bf{2.2} & 2.65	&2.85		\\
&std	&0.27	&0.34	&0.39 & 0.51	\\
\hline
\multicolumn{6}{|c|}{Missing ratio: $0.75$}\\
\hline
\multirow{2}{*}{RIGID}& mean	&1.61	&2.57 & 3.17	&\bf{3.58}		\\
&std	&0.27	&0.44	&0.36 &0.53	\\
\hline
\multirow{2}{*}{RT}& mean	&\bf{1.38}	&\bf{2.37} & \bf{3.16}	&3.77		\\
&std	&0.24 &0.26	&0.3	&0.51	\\
\hline
\end{tabular}}
\scriptsize
\resizebox{6.075cm}{!}{%
\begin{tabular}{|c|c|c|c|c|c|}
\hline 
\multicolumn{6}{|c|}{\boldmath{$n=1000, d=500$}}\\
\hline
\multicolumn{6}{|c|}{Missing ratio: $0.25$}\\
\hline
\multicolumn{2}{|c|}{Sparsity (s)}&$0.2$&$0.4$&$0.6$&$0.8$\\
\hline
\multirow{2}{*}{RIGID}& mean	&3.37	&4.53 &\bf{4.79}	&\bf{5.88}		\\
&std	&0.50	&0.72	&0.75 & 0.82	\\
\hline
\multirow{2}{*}{RT}& mean	&\bf{2.84}	&\bf{4.36} & 5.15	&6.52		\\
&std	&0.31	&0.59	&0.70 & 0.83	\\
\hline
\multicolumn{6}{|c|}{Missing ratio: $0.5$}\\
\hline
\multirow{2}{*}{RIGID}& mean	&4.24	&\bf{5.62} & \bf{6.38}	&\bf{7.15}		\\
&std	&0.54	&0.74	&0.85 & 1.01	\\
\hline
\multirow{2}{*}{RT}& mean	&\bf{3.38}	&5.81 & 6.97	&7.78		\\
&std	&0.45	&0.59	&0.73 & 0.85	\\
\hline
\multicolumn{6}{|c|}{Missing ratio: $0.75$}\\
\hline
\multirow{2}{*}{RIGID}& mean	&4.06	&\bf{6.33} & \bf{7.21}	&\bf{8.05}	\\
&std	&0.23	&0.45 & 0.55	&0.69	\\
\hline
\multirow{2}{*}{RT}& mean	&\bf{4.02}	&6.74 & 7.89	&8.91		\\
&std	&0.28	&0.34 &0.56	&0.53	\\
\hline
\end{tabular}
}
\vspace{-.3cm}
\end{table}

We also perform an extensive set of experiments on real data. For this purpose we consider ten datasets from the UCI machine learning repository. It is important to note that none of these datasets pass the multi-normality test, and we deliberately do this to stress-test RIGID for a general situation, where the multi-normality assumption does not necessarily hold. Panel (d) of Figure \ref{figSemi} shows a comparison of RIGID's performance against other competing techniques in an MCAR setting with 30\% missing rate. We consistently see RIGID among the top performing frameworks both in terms of the average RMSE, and the average mean-absolute error (MAE). The reader is referred to \S~\ref{app:MoreExp} of the Supplementary Material for additional experiments and data/implementation details.

\section{Concluding Remarks}\label{conclude}
This paper presents a framework to use robustness as a way of handling the missing data in linear regression. Our framework ultimately reduces to a unified convex formulation, where the linear fit can be directly acquired from an optimization problem. As a first method of its kind, we focus on linear models, and datasets with elliptical distributions (narrowed down to multi-normal). We believe that the proposed robust formulation can be extended to more sophisticated models, beyond linear regression. Clearly, such extension might limit the access to well-understood conditional distributions, however, the performance boost observed in linear models, encourages one to try more sophisticated predictive schemes and data models. Also, while elliptical distributions cover many interesting data distributions, an extension of the technique to more general multi-modal distributions remains open. 

Theoretically, in this paper we tried to address some of the most immediate and important aspects of the work, such as a scalable solver, understanding of the risk, and the estimation of input parameters. As stated in Theorem \ref{th:LclosedForm}, the structure of the data covariance matrix, the missing patterns and the missing ratio for each pattern play key roles in controlling the behavior of the robust risk.
Presenting results related to the estimation error of the true model (the difference between $\bbeta_0$ and the RIGID Solution) is beyond the current load of the paper, and will be presented in near future as a separate contribution. Moreover, because of the diversity of the input parameters, a rigorous characterization of the worst-case risk, remains an open problem. 

Finally, a generalization of RIGID would be to consider a different $\gamma$ for each missing pattern to more flexibly control the width of the uncertainty regions, $\mathcal{R}_i$. This flexibility comes at the expense of introducing many free parameters, especially when the number of missing patterns is large (e.g., in an MCAR setting). Exploring methods which exploit such flexible formulations without blowing up the problem with redundant hyper-parameters  is another open direction of research. 


\section{Appendix}
\subsection{Proof of Theorem \ref{mainProxTh}}\label{th1proof}
This section presents a step-by-step proof of Theorem \ref{mainProxTh}. To avoid disrupting the flow, the proofs of all the key lemmas are moved to the end of this section as separate subsections.   

Throughout the proof consider $z_1'\in\R$ and $\vz_2'\in\R^p$, $\lambda> 0$ and $\gamma> 0$ given. The goal is acquiring closed-form expressions for the functions $\xi_1$ and $\bxi_2$ defined as
\begin{align}\label{proxBothVars}
 &(\xi_1(\vz'),~ \bxi_2(\vz') )= \argmin_{z_1\in\R, \vz_2\in\R^p}  \frac{\lambda}{2}(|z_1|+\gamma\|\vz_2\|)^2 \!+\! \frac{1}{2 }(z_1-z_1')^2 + \frac{1}{2}\left\|\vz_2 - \vz_2' \right\|^2\!\!\!,
\end{align}
where $\vz' = \begin{bmatrix} z_1'\\ \notag \vz_2'\end{bmatrix}$. In this formulation $\xi_1\in \R$ and $\bxi_2\in\R^p$ are basically the components of the minimizer corresponding to $z_1$ and $\vz_2$, respectively. For $w\geq 0$ define the function
\[\psi(w;z_1',\lambda,\gamma ) = \min_z ~ \frac{\lambda}{2}(|z|+\gamma w)^2 + \frac{1}{2}(z-z_1')^2,
\]
for which the following lemma characterizes the underlying minimizer.
\begin{lemma}\label{lemscalar}
Given a scalar $c\geq 0$, consider the function $\phi_{c}(z) = \frac{1}{2}(|z|+c)^2$. Then for all $\lambda \geq  0$:
\begin{align*} \prox_{\lambda \phi_{c}}(z') = \argmin_z ~ \frac{\lambda}{2}(|z|+c)^2 + \frac{1}{2 }(z-z')^2 = \left\{ \begin{array}{lc} \frac{z'-\sign(z')\lambda c}{1+\lambda} & |z'|>\lambda c\\[.1cm]0 &  |z'| \leq \lambda c \end{array}\right..
\end{align*}
\end{lemma}
\noindent This lemma would provide an explicit form for $\psi(w;z_1',\lambda,\gamma )$. Considering $\phi_{\gamma w}(z) = \frac{1}{2}(|z|+\gamma w)^2$, by plugging in $\prox_{\lambda \phi_{\gamma w}}(z_1')$ in the proximal objective we get: 
\begin{align}\notag 
       \psi(w;z_1',\lambda,\gamma ) &= \frac{\lambda}{2}(|\prox_{\lambda \phi_{\gamma w}}(z_1') |+\gamma w)^2 + \frac{1}{2}(\prox_{\lambda \phi_{\gamma w}}(z_1') -z_1')^2 \\ & =\left\{\begin{array}{lc}\frac{\lambda}{2(1+\lambda)}\left(\gamma w + |z_1'| \right)^2 & 0\leq w <\frac{|z_1'|}{\lambda\gamma }\\ \frac{\lambda}{2}\gamma^2 w^2 + \frac{1}{2}z_1'^2 & w\geq \frac{|z_1'|}{\lambda\gamma } \end{array} \right.. \label{psiEq}
\end{align}
Returning to the minimization in \eqref{proxBothVars}, we can now simplify the underlying program as
\begin{align*}
    &\min_{z_1\in\R, \vz_2\in\R^p}  ~\frac{\lambda}{2}(|z_1|+\gamma\|\vz_2\|)^2 + \frac{1}{2 }(z_1-z_1')^2 + \frac{1}{2}\left\|\vz_2 - \vz_2' \right\|^2 \\& = \min_{\vz_2\in\R^p}\frac{1}{2}\left\|\vz_2 - \vz_2' \right\|^2 \!+\! \min_{z_1\in\R} \frac{\lambda}{2}(|z_1|+\gamma\|\vz_2\|)^2 \!+\! \frac{1}{2 }(z_1-z_1')^2\\ & = \min_{\vz_2\in\R^p}\frac{1}{2}\left\|\vz_2 - \vz_2' \right\|^2 + \psi(\|\vz_2\|; z_1',\lambda,\gamma),
\end{align*}
which simply reveals that $\bxi_2(\vz')$ is the proximal operator of the function $\psi(\|\vz_2\|; z_1',\lambda,\gamma)$. To calculate $\bxi_2(\vz')$, we borrow the following lemma from \cite{beck2017first}: 
\begin{lemma}[Theorem 6.18 in \cite{beck2017first}]\label{lemmabeck}
Let $f:\mathbb{E}\to\R$ be given by $f(\vu) = g(\|\vu\|)$,
where $g: \R\to (-\infty, +\infty]$  is a proper closed and convex function satisfying $\textup{\texttt{dom}}(g) \subseteq [0, \infty)$. Then
\[\prox_f(\vu') = \left\{\begin{array}{lc} \prox_g(\|\vu'\|)\frac{\vu'}{\|\vu'\|} & \vu' \neq \boldsymbol{0} \\ \{\vv\in\mathbb{E}: \|\vv\| = \prox_g(0)\} & \vu'=\boldsymbol{0} \end{array} \right..
\]
\end{lemma}
\noindent Now consider the extended-value function 
\begin{equation}\label{psiExtended}
\psi^E(w;z_1',\lambda,\gamma) = \left\{\begin{array}{lc} \psi(w;z_1',\lambda,\gamma) & w\geq 0\\ +\infty & w<0 \end{array}\right..
\end{equation}
Clearly, $\psi(\|\vz_2\|; z_1',\lambda,\gamma)$ and $\psi^E(\|\vz_2\|; z_1',\lambda,\gamma)$ offer identical proximal operators, i.e., 
\begin{align*}
   \bxi_2(\vz') &= \argmin_{\vz_2\in\R^p}~\frac{1}{2}\left\|\vz_2 - \vz_2' \right\|^2 + \psi(\|\vz_2\|; z_1',\lambda,\gamma) \\& = \argmin_{\vz_2\in\R^p}~\frac{1}{2}\left\|\vz_2 - \vz_2' \right\|^2 + \psi^E(\|\vz_2\|; z_1',\lambda,\gamma).
\end{align*}
On the other hand $\psi^E(w;z_1',\lambda,\gamma)$ meets all the conditions required for the function $g(.)$ in Lemma \ref{lemmabeck}. Hence, we can apply the lemma and claim that
\begin{align}\label{afterApplyingBeck}
   \bxi_2(\vz') \!=\!  \left\{\begin{array}{lc} \prox_{\psi^E(.;z_1',\lambda,\gamma)}(\|\vz_2'\|)\frac{\vz_2'}{\|\vz_2'\|} & \vz_2' \neq \boldsymbol{0} \\ \{\vv: \|\vv\|\! =\! \prox_{\psi^E(.;z_1',\lambda,\gamma)}(0)\} & \vz_2'=\boldsymbol{0} \end{array} \right.\!\!\!\!.
\end{align}
The following lemma presents a closed-form expression for the proximal operator of $\psi^E(w;z_1',\lambda,\gamma)$:

\begin{lemma}\label{lemproxg}
Consider $x\in \R$, $\lambda>0$, $\gamma>0$ given, and the extended-value function $\psi^E(w;x,\lambda,\gamma)$ as characterized in \eqref{psiExtended} and \eqref{psiEq}. Then
\begin{align} \label{proxg}
\prox_{\psi^E(.;x,\lambda,\gamma)}(w') = \left\{ \begin{array}{lc}
\frac{w'}{1+\lambda\gamma^2} & w'\geq \frac{(1+\lambda\gamma^2)|x|}{\lambda\gamma}\\
 \frac{(\lambda+1)w'-\lambda\gamma |x|}{1+\lambda+\lambda\gamma^2}& 
 \frac{\lambda\gamma |x|}{\lambda+1}\leq w' < \frac{(1+\lambda\gamma^2)|x|}{\lambda\gamma}
 \\
 0 & w'<\frac{\lambda\gamma |x|}{\lambda+1} \end{array}\right. .
\end{align}
\end{lemma}
\noindent After applying Lemma \ref{lemproxg} to \eqref{afterApplyingBeck}, we see that for $\vz' = \begin{bmatrix} z_1'\\ \vz_2'\end{bmatrix}$:
\begin{align}\label{afterApplyingBeckEv}
 \bxi_2(\vz')\! =\! \left\{\!\!\! \begin{array}{lc}
\frac{1}{1+\lambda\gamma^2}\vz_2' & \|\vz_2'\|\geq \frac{(1+\lambda\gamma^2)|z_1'|}{\lambda\gamma}\\[.2cm]
 \frac{(\lambda+1)\|\vz_2'\|-\lambda\gamma |z_1'|}{(1+\lambda+\lambda\gamma^2)\|\vz_2'\|} \vz_2'& 
 \frac{\lambda\gamma |z_1'|}{\lambda+1}\leq \|\vz_2'\| < \frac{(1+\lambda\gamma^2)|z_1'|}{\lambda\gamma}
 \\[.2cm]
 \boldsymbol{0} & \|\vz_2'\|<\frac{\lambda\gamma |z_1'|}{\lambda+1} \end{array}\right.\!\!\!\!.
\end{align}
Now that $\bxi_2(\vz')$ is obtained, we plug its value in \eqref{proxBothVars} and perform the minimization in terms of $z_1$ to acquire $\xi_1(\vz')$. Doing this would give
\begin{align}
    \notag \xi_1(\vz') &= \argmin_{z\in\R} \frac{\lambda}{2}(|z|\!+\!\gamma\|\bxi_2(\vz')\|)^2 \!+\! \frac{1}{2 }(z_-z_1')^2 \!+\! \frac{1}{2}\left\|\bxi_2(\vz') \!-\! \vz_2' \right\|^2\\ \notag &= \argmin_{z\in\R} \frac{\lambda}{2}(|z|\!+\!\gamma\|\bxi_2(\vz')\|)^2 + \frac{1}{2 }(z_-z_1')^2\\&= \left\{ \begin{array}{lc} \frac{z_1'-\sign(z_1')\lambda \gamma \|\bxi_2(\vz')\|}{1+\lambda} & |z_1'|>\lambda \gamma  \|\bxi_2(\vz')\|\\[.1cm]0 &  |z_1'| \leq \lambda \gamma \|\bxi_2(\vz') \| \end{array}\right.,\label{xi1Exp}
\end{align}
where the last equality is thanks to Lemma \ref{lemscalar}. We can further simplify the expression in \eqref{xi1Exp}, as detailed below:
\begin{itemize}
    \item[--] If  $z_1'=0$, or $\|\vz_2'\|\geq \frac{(1+\lambda\gamma^2)|z_1'|}{\lambda\gamma}$, then from \eqref{afterApplyingBeckEv} $\|\bxi_2(\vz')\| = \frac{1}{1+\lambda\gamma^2}\|\vz_2'\|$, which together give $\lambda\gamma \|\bxi_2(\vz')\| \geq |z_1'|$, and as a result of \eqref{xi1Exp}, $\xi_1(\vz')=0$.
    \item[--] If  $ \frac{\lambda\gamma |z_1'|}{\lambda+1}\leq \|\vz_2'\| < \frac{(1+\lambda\gamma^2)|z_1'|}{\lambda\gamma}$, then from \eqref{afterApplyingBeckEv} 
    \begin{align*}\|\bxi_2(\vz')\| = \frac{(\lambda+1)\|\vz_2'\|-\lambda\gamma |z_1'|}{(1+\lambda+\lambda\gamma^2)}< \frac{(\lambda+1)\frac{(1+\lambda\gamma^2)|z_1'|}{\lambda\gamma}-\lambda\gamma |z_1'|}{(1+\lambda+\lambda\gamma^2)} = \frac{|z_1'|}{\lambda\gamma}, 
    \end{align*}
    which as a result of \eqref{xi1Exp}, gives 
    \begin{align*}\xi_1(\vz')&= \frac{z_1'-\sign(z_1')\lambda \gamma \frac{(\lambda+1)\|\vz_2'\|-\lambda\gamma |z_1'|}{(1+\lambda+\lambda\gamma^2)}}{1+\lambda} \\&=\frac{(1+\lambda+\lambda\gamma^2+\lambda^2\gamma^2)z_1'-\lambda \gamma (\lambda+1)\sign(z_1')\|\vz_2'\|}{(1+\lambda)(1+\lambda+\lambda\gamma^2)}\\ & = \frac{(1+\lambda\gamma^2)z_1' - \lambda\gamma \sign(z_1')\|\vz_2'\|}{1+\lambda+\lambda\gamma^2}.
    \end{align*}
    \item[--] Finally, if $z_1'\neq 0$ and $\|\vz_2'\|<\frac{\lambda\gamma |z_1'|}{\lambda+1}$, then from \eqref{afterApplyingBeckEv} $\|\bxi_2(\vz')\| = 0$, and based on \eqref{xi1Exp} $\xi_1(\vz')=\frac{z_1'}{1+\lambda}$.
\end{itemize}
For the extreme case $\lambda=0$, \eqref{proxBothVars} trivially returns $\xi_1(\vz')=z_1'$ and $\bxi_2(\vz')=\vz_2'$. For the other extreme case $\gamma=0$, again \eqref{proxBothVars} returns $\xi_1(\vz')=\frac{z_1'}{1+\lambda}$ and $\bxi_2(\vz')=\vz_2'$. These results are consistent with the formulation advertised in the theorem, and the proof of Theorem \ref{mainProxTh} is complete.

\subsubsection{Proof of Lemma \ref{lemscalar}}
Given $c\geq 0$ and $z'$, define the function 
\begin{align*}p(z;z') &= \frac{\lambda}{2}(|z|+c)^2 + \frac{1}{2}(z-z')^2 \\& = \frac{\lambda}{2}z^2 + \frac{\lambda}{2}c^2 +\frac{1}{2}(z-z')^2 + \lambda c|z|.
\end{align*}
The following scenarios characterize the minimizer of $p(z;z')$ with respect to $z$, denoted by $z^*$:
\begin{itemize}
    \item[--]  Assuming $z^*\neq 0$, taking a derivative of $p(z;z')$ with respect to $z$ and setting it to zero yields $(\lambda+1)z^* - z' + \lambda c\sign(z^*)=0$. This implies that $z^* = (z'-\lambda c)/(1+\lambda)$ if $z'>\lambda c\geq 0$, and $z^* = (z'+\lambda c)/(1+\lambda)$ if $z'<-\lambda c\leq 0$.
    \item[--]  To have $z^*=0$, it is required that $0\in\partial p(0;z')$, where $\partial p(0;z')$ is the subdifferential of $p(z;z')$ evaluated at $z=0$. Basic derivation yields $\partial p(0;z') = \{-z' + \lambda c g: |g|\leq 1\}$. Hence, having $0\in\partial p(0;z')$ is possible when $|z'|\leq \lambda c$. 
\end{itemize}
These results are consistent with the formulation advertised in Lemma \ref{lemscalar}.

\subsubsection{Proof of Lemma \ref{lemproxg}}
It is easy to verify that $\psi^E$ can be more compactly represented as
\begin{align*}
\psi^E(w;x,\lambda,\gamma) = 
\frac{1}{2}x^2+ \frac{1}{2}\lambda\gamma^2 w^2 - \frac{1}{4(\lambda+1)}(\lambda\gamma w-|x|)^2  + \frac{1}{4(\lambda+1)}(\lambda\gamma w-|x|)|\lambda\gamma w-|x|~\!|
\end{align*}
for $w\geq 0$, and $\psi^E(w;x,\lambda,\gamma)=+\infty$ for $w<0$. We define the proximal objective
\begin{align*}P(w;w') = \frac{1}{2}(w-w')^2 + \frac{1}{2}x^2+ \frac{1}{2}\lambda\gamma^2 w^2 - \frac{1}{4(\lambda+1)}(\lambda\gamma w-|x|)^2 + \frac{1}{4(\lambda+1)}(\lambda\gamma w-|x|)|\lambda\gamma w-|x|~\!|.
\end{align*}
Minimizing $P(w,w')$ with respect to $w$ and over the domain $w\geq 0$ gives $\prox_{\psi^E(.;x,\lambda,\gamma)}(w')$. For this purpose, notice that despite the the absolute-value term, $P(w;w')$ is differentiable for all $w> 0$ (also right-differentiable at $w=0$), and 
\begin{equation}\label{pzzp}
\frac{\partial}{\partial w}P(w;w') =w - w' + \lambda\gamma^2 w - \frac{\lambda \gamma}{\lambda+1}\min(\lambda\gamma w-|x|,0).
\end{equation}
The following scenarios characterize the minimizer of $P(w;w')$, denoted by $w^*$:
\begin{itemize}
    \item[--] Assuming $\lambda\gamma w^*\geq |x|$, setting \eqref{pzzp} to zero gives $w^* = w'/(1+\lambda\gamma^2)$. In other words, if  $w'\geq (1+\lambda\gamma^2)|x|/(\lambda\gamma)$, then $w^* = w'/(1+\lambda\gamma^2)$.
    \item[--] Assuming $0<\lambda\gamma w^*\leq |x|$, setting \eqref{pzzp} to zero gives $w^* = ((\lambda+1)w'-\lambda\gamma |x|)/(1+\lambda+\lambda\gamma^2)$. In other words, if  $\lambda\gamma |x|/(\lambda+1)\leq w' < (1+\lambda\gamma^2)|x|/(\lambda\gamma)$, then $w^* = ((\lambda+1)w'-\lambda\gamma |x|)/(1+\lambda+\lambda\gamma^2)$.
    \item[--] To have $w^*=0$, constrained optimality condition requires that for all $v\geq 0$: 
    \[v\frac{\partial}{\partial w}P(w;w')|_{w=0}\geq 0,
    \]
    which implies that $v(-w' + \lambda\gamma |x|/(\lambda+1))\geq 0$, or $ w'\leq \lambda\gamma |x|/(\lambda+1)$. 
\end{itemize}
These results are consistent with the formulation in \eqref{proxg}, and the proof is complete. 

\subsection{Proof of Theorem \ref{th:robustRiskSparse}}
\label{th3proof}Before we start the proof we would like to make a note about the positivity of the matrices that appear in the robust risk \eqref{th:eqLBeta}. With no loss of generality, for $\bSigma\succ \boldsymbol{0}$ consider the following structure
\begin{equation}\label{eqCovSinglePattern}\bSigma = \begin{bmatrix} \bSigma_{\AAA}&\bSigma_{\AM}\\ \bSigma_{\MA} & \bSigma_{\MM}\end{bmatrix}, 
\end{equation}
where $\eM$ is a missing pattern and $\mathcal{A} = \eM^c$. Since $\bSigma\succ\boldsymbol{0}$, we immediately get $\bSigma_{\AAA}\succ \boldsymbol{0}$, because $\bSigma\succ \boldsymbol{0}$ and for any nonzero vector $\vv$, we have 
\[\vv^\top\bSigma_{\AAA}\vv = \begin{bmatrix}\vv\\ \boldsymbol{0} \end{bmatrix}^\top \bSigma\begin{bmatrix}\vv\\ \boldsymbol{0} \end{bmatrix}>0.
\]
By the properties of the Schur complement we know that $\bSigma\succ \boldsymbol{0}$ implies $\bar\bSigma_{\eM}={\bSigma\slash \bSigma}_{\AAA}\succ\boldsymbol{0}$. Furthermore, for $\bar \bSigma_{\eM}^E$ defined as \eqref{eqSigmaE} we get
\[
\bSigma - \bar \bSigma_{\eM}^E =  \begin{bmatrix} \bSigma_{\AAA}&\bSigma_{\AM}\\ \bSigma_{\MA} & \bSigma_{\MA}\bSigma_{\AAA}^{-1}\bSigma_{\AM}\end{bmatrix}, 
\]
which is known to be positive semi-definite, thanks to $\bSigma_{\AAA}\succ \boldsymbol{0}$ (see 14.27 in \cite{seber2008matrix}). In summary, in the robust risk \eqref{th:eqLBeta}, both matrices $\bar \bSigma_{\eM_j}^E$ and $\bSigma - \bar \bSigma_{\eM_j}^E$ are positive semi-definite. While the Schure complement matrix $\bar \bSigma_{\eM_j}$ is strictly positive definite, $\bar \bSigma_{\eM_j}^E$ which is basically its zero-padded version acquired through \eqref{eqSigmaE} becomes positive semi-definite. \\

\noindent \textbf{Part (a).} When $\gamma=0$, the robust risk $\mathcal{L}(\bbeta;\gamma)$ in \eqref{th:eqLBeta} reduces to
\begin{align}\notag 
    \mathcal{L} (\bbeta;0) &= \sigma^2 + \sum_{j=1}^M \pi_j \left( \left\|\bbeta-\bbeta_0\right\|^2_{\bSigma - {\bar \bSigma_{\eM_j}^E}} + \left\|\bbeta_0\right\|^2_{ {\bar \bSigma_{\eM_j}^E}} \right)\\ & = \label{eqgamma0} \sigma^2 + \left(\bbeta-\bbeta_0\right)^\top\left(\sum_{j=1}^M\pi_j\left(\bSigma - {\bar \bSigma_{\eM_j}^E} \right) \right)\left(\bbeta-\bbeta_0\right) + \sum_{j=1}^M \pi_j\left\|\bbeta_0\right\|^2_{ {\bar \bSigma_{\eM_j}^E}}, 
\end{align}
which becomes a strongly convex function of $\bbeta$ (presenting a unique minimizer), if and only if 
\[\sum_{j=1}^M\pi_j\left(\bSigma - {\bar \bSigma_{\eM_j}^E} \right)\succ \boldsymbol{0}. 
\]

\noindent \textbf{Part (b).} One observes that the robust risk $\mathcal{L}(\bbeta;\gamma)$ in \eqref{th:eqLBeta} can be written as 
\begin{align}\notag
    \mathcal{L}(\bbeta;\gamma ) = &\underbrace{\sum_{j=1}^M\pi_j \left(  \left( \sigma^2 + \left\|\bbeta-\bbeta_0\right\|^2_{\bSigma - {\bar \bSigma_{\eM_j}^E}} + \left\|\bbeta_0\right\|^2_{{\bar \bSigma_{\eM_j}^E}}\! \right)^{\frac{1}{2}} + \sqrt{\frac{2}{\pi}}\gamma \left\|\bbeta \right\|_{{\bar \bSigma_{\eM_j}^E}} \right)^2}_{\ell_1(\bbeta)} +\gamma^2\underbrace{\sum_{j=1}^M\pi_j \left(1-\frac{2}{\pi}\right)\left\|\bbeta\right\|_{\bar \bSigma_{\eM_j}^E}^2}_{\ell_2(\bbeta)}.\label{eqLbetagamma}
\end{align}
If we show that $\mathcal{L}(\bbeta;\gamma )$ is strictly convex, the uniqueness of the minimizer immediately follows. For this purpose, it suffices to show that if $\eM\neq\emptyset$, and $\gamma>0$, the following function is strictly convex:
\begin{equation} \label{def_lb}
    \ell(\bbeta) = \left(  \left( 1 + \left\|\bbeta-\bbeta_0\right\|^2_{\bSigma - {\bar \bSigma_{\eM}^E}}  \right)^{\frac{1}{2}} + \gamma \left\|\bbeta \right\|_{{\bar \bSigma_{\eM}^E}} \right)^2.
\end{equation}
This is because  $ \ell_1(\bbeta)$ part of $\mathcal{L}(\bbeta;\gamma )$ is a weighted summation of such functions, and thanks to the convexity of $ \ell_2(\bbeta)$, the strict convexity of $\ell(\bbeta)$ is enough for $\mathcal{L}(\bbeta;\gamma )$ to become strictly convex. This would be the focus of the remainder of the proof.

Since $\ell(\bbeta)$ is continuous, it suffices to show the strict midpoint convexity. More specifically showing that for any pair of points $\bbeta_1$ and $\bbeta_2$, where $\bbeta_1\neq \bbeta_2$:
\[\Delta(\bbeta_1,\bbeta_2)\triangleq \ell(\bbeta_1)+\ell(\bbeta_2) - 2\ell(\frac{\bbeta_1+\bbeta_2}{2})>0. 
\]
Expanding $\Delta(\bbeta_1,\bbeta_2)$ based on the formulation of $\ell(\bbeta)$ gives
\begin{align}
\Delta(\bbeta_1,\bbeta_2) &=  \left\|\bbeta_1-\bbeta_0\right\|^2_{\bSigma - {\bar \bSigma_{\eM}^E}} + \gamma^2 \left\|\bbeta_1 \right\|_{{\bar \bSigma_{\eM}^E}}^2 + 2\gamma \left\|\bbeta_1 \right\|_{{\bar \bSigma_{\eM}^E}} \left( 1 + \left\|\bbeta_1-\bbeta_0\right\|^2_{\bSigma - {\bar \bSigma_{\eM}^E}}  \right)^{\frac{1}{2}}\nonumber\\
&~~~~~ +  \left\|\bbeta_2-\bbeta_0\right\|^2_{\bSigma - {\bar \bSigma_{\eM}^E}} + \gamma^2 \left\|\bbeta_2 \right\|_{{\bar \bSigma_{\eM}^E}}^2 + 2\gamma \left\|\bbeta_2 \right\|_{{\bar \bSigma_{\eM}^E}} \left( 1 + \left\|\bbeta_2-\bbeta_0\right\|^2_{\bSigma - {\bar \bSigma_{\eM}^E}}  \right)^{\frac{1}{2}}\nonumber\\ \notag 
&~~~~~ -2 \left\|\tfrac{\bbeta_1+\bbeta_2}{2}-\bbeta_0\right\|^2_{\bSigma - {\bar \bSigma_{\eM}^E}} - \frac{\gamma^2}{2} \left\|\bbeta_1+\bbeta_2 \right\|_{{\bar \bSigma_{\eM}^E}}^2  -2 \gamma \left\|\bbeta_1+\bbeta_2\right\|_{{\bar \bSigma_{\eM}^E}} \left( 1 + \left\|\tfrac{\bbeta_1+\bbeta_2}{2}-\bbeta_0\right\|^2_{\bSigma - {\bar \bSigma_{\eM}^E}}  \right)^{\frac{1}{2}}\nonumber\\
& = \frac12 \left(\left\|\bbeta_1-\bbeta_0\right\|^2_{\bSigma - {\bar \bSigma_{\eM}^E}} + \gamma^2 \left\|\bbeta_1 \right\|_{{\bar \bSigma_{\eM}^E}}^2+  \left\|\bbeta_2-\bbeta_0\right\|^2_{\bSigma - {\bar \bSigma_{\eM}^E}} + \gamma^2 \left\|\bbeta_2 \right\|_{{\bar \bSigma_{\eM}^E}}^2 \right)\nonumber\\
&~~~~~ -(\bbeta_1-\bbeta_0)^\top (\bSigma - \bar \bSigma_{\eM}^E) (\bbeta_2-\bbeta_0)
- \gamma^2 \bbeta_1^\top \bar \bSigma_{\eM}^E \bbeta_2+2\gamma C(\bbeta_1,\bbeta_2),\label{mid_strict}
\end{align}
where
\begin{align}\notag 
C(\bbeta_1,\bbeta_2) &= \left\|\bbeta_1 \right\|_{{\bar \bSigma_{\eM}^E}} \left( 1 + \left\|\bbeta_1-\bbeta_0\right\|^2_{\bSigma - {\bar \bSigma_{\eM}^E}}  \right)^{\frac{1}{2}}+
\left\|\bbeta_2 \right\|_{{\bar \bSigma_{\eM}^E}} \left( 1 + \left\|\bbeta_2-\bbeta_0\right\|^2_{\bSigma - {\bar \bSigma_{\eM}^E}}  \right)^{\frac{1}{2}} \\ \notag &~~~ -
\left\|\bbeta_1+\bbeta_2\right\|_{{\bar \bSigma_{\eM}^E}} \left( 1 + \left\|\tfrac{\bbeta_1+\bbeta_2}{2}-\bbeta_0\right\|^2_{\bSigma - {\bar \bSigma_{\eM}^E}}  \right)^{\frac{1}{2}}\\ & =  2\left\|\bbeta_1 \right\|_{{\bar \bSigma_{\eM}^E}} \left\|\tilde \bbeta_1 \right\|_{\bSigma^\dagger}+
2\left\|\bbeta_2 \right\|_{{\bar \bSigma_{\eM}^E}} \left\|\tilde \bbeta_2 \right\|_{\bSigma^\dagger}-
\left\|\bbeta_1+\bbeta_2\right\|_{{\bar \bSigma_{\eM}^E}} \left\|\tilde \bbeta_1 + \tilde \bbeta_2 \right\|_{\bSigma^\dagger}, \label{def_C}
\end{align}
for which we used the change of variables 
\[
\tilde \bbeta_i = \frac12 \begin{bmatrix}
1  \\
\bbeta_i - \bbeta_0
\end{bmatrix} \qquad i=1,2, \qquad \mbox{and} \qquad
\bSigma^\dagger = \begin{bmatrix}
1 & \boldsymbol{0} \\
\boldsymbol{0} & \bSigma - {\bar \bSigma_{\eM}^E}
\end{bmatrix}.
\]
Noting that $\bSigma^\dagger$ is also a PSD matrix, by the triangle inequality, we have 
\begin{align}\nonumber 
\left\|\bbeta_1+\bbeta_2\right\|_{{\bar \bSigma_{\eM}^E}}  \left\|\tilde \bbeta_1 + \tilde \bbeta_2 \right\|_{\bSigma^\dagger}
&\le \left(\left\|\bbeta_1\right\|_{{\bar \bSigma_{\eM}^E}}+\left\|\bbeta_2\right\|_{{\bar \bSigma_{\eM}^E}} \right) \left\|\tilde \bbeta_1 + \tilde \bbeta_2 \right\|_{\bSigma^\dagger}\nonumber\\
&= \left\|2 \left\|\bbeta_1\right\|_{{\bar \bSigma_{\eM}^E}} \tilde \bbeta_1 + 2 \left\|\bbeta_2\right\|_{{\bar \bSigma_{\eM}^E}}\tilde \bbeta_2 
+ \left(\left\|\bbeta_1\right\|_{{\bar \bSigma_{\eM}^E}}-\left\|\bbeta_2\right\|_{{\bar \bSigma_{\eM}^E}}\right)\left(\tilde \bbeta_2 -\tilde \bbeta_1 \right) \right\|_{\bSigma^\dagger}\nonumber\\
&\le \left\|2 \left\|\bbeta_1\right\|_{{\bar \bSigma_{\eM}^E}} \tilde \bbeta_1 + 2 \left\|\bbeta_2\right\|_{{\bar \bSigma_{\eM}^E}}\tilde \bbeta_2 \right\|_{\bSigma^\dagger} +\left|\left\|\bbeta_1\right\|_{{\bar \bSigma_{\eM}^E}}-\left\|\bbeta_2\right\|_{{\bar \bSigma_{\eM}^E}} \right| \left\|
 \left(\tilde \bbeta_2 -\tilde \bbeta_1 \right) \right\|_{\bSigma^\dagger} \nonumber \\
&\le 2 \left\|\bbeta_1\right\|_{{\bar \bSigma_{\eM}^E}} \left\|\tilde \bbeta_1 \right\|_{\bSigma^\dagger}+ 2 \left\|\bbeta_2\right\|_{{\bar \bSigma_{\eM}^E}} \left\|\tilde \bbeta_2\right\|_{\bSigma^\dagger} 
+ \left\|\bbeta_1-\bbeta_2\right\|_{{\bar \bSigma_{\eM}^E}}\left\|\tilde \bbeta_2 -\tilde \bbeta_1 \right\|_{\bSigma^\dagger}\nonumber \\
&= 2 \left\|\bbeta_1\right\|_{{\bar \bSigma_{\eM}^E}} \left\|\tilde \bbeta_1 \right\|_{\bSigma^\dagger}+ 2 \left\|\bbeta_2\right\|_{{\bar \bSigma_{\eM}^E}} \left\|\tilde \bbeta_2\right\|_{\bSigma^\dagger} 
+ \frac12 \left\|\bbeta_1-\bbeta_2\right\|_{{\bar \bSigma_{\eM}^E}}\left\| \bbeta_2 - \bbeta_1 \right\|_{\bSigma - {\bar \bSigma_{\eM}^E}}\!.\label{ineq1}
\end{align}
Combining the above inequality with \eqref{def_C} and \eqref{mid_strict}, we get
\begin{align*}
\Delta(\bbeta_1,\bbeta_2)
&\ge \frac12 \left(\left\|\bbeta_1-\bbeta_0\right\|^2_{\bSigma - {\bar \bSigma_{\eM}^E}} + \gamma^2 \left\|\bbeta_1 \right\|_{{\bar \bSigma_{\eM}^E}}^2+  \left\|\bbeta_2-\bbeta_0\right\|^2_{\bSigma - {\bar \bSigma_{\eM}^E}} + \gamma^2 \left\|\bbeta_2 \right\|_{{\bar \bSigma_{\eM}^E}}^2 \right)\nonumber\\
& -(\bbeta_1-\bbeta_0)^\top (\bSigma - \bar \bSigma_{\eM}^E) (\bbeta_2-\bbeta_0)
- \gamma^2 \bbeta_1^\top \bar \bSigma_{\eM}^E \bbeta_2 -\gamma \left\|\bbeta_1-\bbeta_2\right\|_{{\bar \bSigma_{\eM}^E}}\left\| \bbeta_2 -\bbeta_1 \right\|_{\bSigma - {\bar \bSigma_{\eM}^E}}\nonumber\\
&=\frac12 \left(\left\| \bbeta_2 -\bbeta_1 \right\|_{\bSigma - {\bar \bSigma_{\eM}^E}}-\gamma \left\|\bbeta_2-\bbeta_1\right\|_{{\bar \bSigma_{\eM}^E}} \right)^2 \ge 0, 
\end{align*}
or in short
\begin{align}\label{strict2}
    \Delta(\bbeta_1,\bbeta_2) \geq \frac12 \left(\left\| \bbeta_2 -\bbeta_1 \right\|_{\bSigma - {\bar \bSigma_{\eM}^E}}-\gamma \left\|\bbeta_2-\bbeta_1\right\|_{{\bar \bSigma_{\eM}^E}} \right)^2 \ge 0. 
\end{align}
 Next, notice that $\bSigma^\dagger$ being PSD admits the representation 
\begin{equation}\label{qcc}
    \bSigma^\dagger = \tilde\mC\tilde \mC^\top, 
\end{equation}
for some matrix $\tilde\mC$ (no necessarily unique). For two nonzero vectors $\tilde \vu$ and $\tilde \vv$, $\|\tilde\vu+\tilde\vv\|_{\bSigma^\dagger} = \|\tilde\vu\|_{\bSigma^\dagger} + \|\tilde\vv\|_{\bSigma^\dagger}$ implies that for some $\tilde\mC$ obeying \eqref{qcc}, $\|\tilde \mC^\top(\tilde\vv+\tilde\vv)\| = \|\tilde \mC^\top\tilde\vu\| + \|\tilde \mC^\top \tilde\vv\|$, which in turn implies that for some $\kappa> 0$, $\tilde \mC^\top\tilde\vu = \kappa \tilde \mC^\top \tilde\vv$ (the condition that the triangle inequality becomes equality). Multiplying both sides of this equation by $\tilde\mC$, we can claim that for two nonzero vectors $\tilde \vu$ and $\tilde \vv$
\begin{equation}\label{suffTriangle}
\|\tilde\vu+\tilde\vv\|_{\bSigma^\dagger} = \|\tilde\vu\|_{\bSigma^\dagger} + \|\tilde\vv\|_{\bSigma^\dagger} \implies \exists \kappa>0: \bSigma^\dagger\tilde\vu = \kappa\bSigma^\dagger\vv.  
\end{equation}
Using this result, we now show that whenever $\bbeta_2\neq \bbeta_1$, the first inequality in \eqref{strict2} becomes strict. To this end, we consider the following two complement cases:
\begin{itemize}
\item \textbf{Case I:} $\bbeta_2\neq \bbeta_1$ and $\|\bbeta_2\|_{{\bar \bSigma_{\eM}^E}} \neq \|\bbeta_1\|_{{\bar \bSigma_{\eM}^E}}.$\\
In this case if the first inequality in \eqref{strict2} becomes an equality, then the second inequality in \eqref{ineq1} should also have been an equality. Since $\left(\left\|\bbeta_1\right\|_{{\bar \bSigma_{\eM}^E}}-\left\|\bbeta_2\right\|_{{\bar \bSigma_{\eM}^E}}\right)\left(\tilde \bbeta_2 -\tilde \bbeta_1 \right)\neq 0$, appealing to \eqref{suffTriangle}, the equality implies that for some $\kappa>0$:
\begin{equation}\label{Qcon}
    \bSigma^\dagger\left( \left\|\bbeta_1\right\|_{{\bar \bSigma_{\eM}^E}} \tilde \bbeta_1 +  \left\|\bbeta_2\right\|_{{\bar \bSigma_{\eM}^E}}\tilde \bbeta_2  \right) = \kappa  
\left(\left\|\bbeta_1\right\|_{{\bar \bSigma_{\eM}^E}}-\left\|\bbeta_2\right\|_{{\bar \bSigma_{\eM}^E}}\right)\bSigma^\dagger \left(\tilde \bbeta_2 -\tilde \bbeta_1 \right). 
\end{equation}
However, since
\[\tilde \bbeta_2 - \tilde \bbeta_1= \frac12 \begin{bmatrix}
0  \\
\bbeta_2 - \bbeta_1\end{bmatrix},
\]
the first component of the left-hand side vector in \eqref{Qcon} is $\|\bbeta_2\|_{{\bar \bSigma_{\eM}^E}} + \|\bbeta_1\|_{{\bar \bSigma_{\eM}^E}}$, while the first component of the right-hand side vector is 0, which is a contradiction. 

\item \textbf{Case II:} $\bbeta_2\neq \bbeta_1$  and $\|\bbeta_2\|_{{\bar \bSigma_{\eM}^E}} = \|\bbeta_1\|_{{\bar \bSigma_{\eM}^E}}$.\\
In this case, since $\bar\bSigma_{\eM}$ is positive definite, and $\tilde\bbeta_1 + \tilde\bbeta_2\neq\boldsymbol{0}$,  the first inequality in \eqref{ineq1} only happens when 
\[\bbeta_{1,\eM} = \kappa \bbeta_{2,\eM}, 
\]
for some $\kappa>0$. This is certainly a contradiction, unless $\bbeta_{1,\eM} = \bbeta_{2,\eM}$. Now suppose $\bbeta_{1,\eM} = \bbeta_{2,\eM}$, then the third inequality in \eqref{ineq1} implies that for some $\kappa'>0$:
\[\bSigma^\dagger\tilde\bbeta_1 = \kappa'\bSigma^\dagger\tilde\bbeta_2,\]
or 
\begin{equation}\label{qbkappap}
    \begin{bmatrix} 1\\ (\bSigma - \bar \bSigma_{\eM}^E)(\bbeta_1-\bbeta_0)\end{bmatrix} = \begin{bmatrix} \kappa'\\ \kappa'(\bSigma - \bar \bSigma_{\eM}^E)(\bbeta_2-\bbeta_0)\end{bmatrix}. 
\end{equation}
Equation \eqref{qbkappap} requires $\kappa' = 1$, and since $\bbeta_{1,\eM} = \bbeta_{2,\eM}$, setting the second blocks equal implies that
\[\bSigma (\bbeta_2 - \bbeta_1) = \boldsymbol{0},
\]
which is again a contradiction since $\bbeta_2\neq \bbeta_1$. 
\end{itemize}


\noindent \textbf{Part (c).} When $\gamma=0$, the robust risk $\mathcal{L}(\bbeta;\gamma)$ in \eqref{th:eqLBeta} reduces to \eqref{eqgamma0}, which is a strongly convex function of $\bbeta$ when the condition in part (a) holds, and trivially attains $\bbeta^\gamma = \bbeta_0$ as the solution. To show the solution continuity and the convergence of the minimizer to $\bbeta_0$ when $\gamma\to 0$, we use the maximum theorem under convexity. A variant of the theorem customized for our problem is stated below:
\begin{lemma}[Theorem 9.17, part 2, in \cite{sundaram1996first}]
\label{lemMaxTh} Suppose that over its domain, $f(\bbeta;\gamma)$ is jointly continuous in $\bbeta$ and $\gamma$, and $\boldsymbol{\mathcal{D}}$ is a compact-valued continuous correspondence on $\gamma$. Let \begin{align*}
\boldsymbol{\mathcal{D}}^*(\gamma) = \argmax_{\bbeta\in \boldsymbol{\mathcal{D}}(\gamma)} ~ f(\bbeta;\gamma).
\end{align*}
If for each $\gamma$ in the domain $f(.;\gamma)$ is strictly concave in $\bbeta$, and $\boldsymbol{\mathcal{D}}(\gamma)$ is a convex set, then $\boldsymbol{\mathcal{D}}^*$ is a continuous function of $\gamma$. 
\end{lemma}
To use this result, consider the closed interval $\Gamma=[0,1]$, and define the set  
\begin{align*}\boldsymbol{\mathcal{D}}_1 &= \left\{\bbeta\in \R^p: \mathcal{L} (\bbeta;0)\leq \mathcal{L} (\boldsymbol{0};1)\right \}\\ &= \left \{\bbeta\in \R^p: \mathcal{L} (\bbeta;0)\leq \sigma^2 + \|\bbeta_0\|^2_{\bSigma}\right \},
\end{align*}
where $\mathcal{L}(\bbeta;0)$ and $\mathcal{L}(\bbeta;1)$ indicate the values of $\mathcal{L}(\bbeta;\gamma)$ when $\gamma=0$ and $\gamma=1$, respectively. The strict convexity of the function $\mathcal{L} (\bbeta; 0)$ guarantees that $\boldsymbol{\mathcal{D}}_1$ is bounded and closed, which together imply its compactness. It is also obvious that $\boldsymbol{\mathcal{D}}_1$ is convex. Define the correspondence 
\[\boldsymbol{\mathcal{D}}(\gamma) = \boldsymbol{\mathcal{D}}_1, ~~~~ \forall \gamma\in\Gamma, \]
which is a compact-valued and continuous correspondence.
Using Lemma \ref{lemMaxTh} immediately implies that 
\begin{align*}
\boldsymbol{\mathcal{D}}^*(\gamma) = \argmax_{\bbeta\in \boldsymbol{\mathcal{D}}_1} ~ -\mathcal{L} (\bbeta;\gamma) = \argmin_{\bbeta\in \boldsymbol{\mathcal{D}}_1} ~ \mathcal{L} (\bbeta;\gamma),
\end{align*}
is a continuous function of $\gamma$ in $\Gamma$, and specifically 
\begin{equation}\label{eqDStar}
 \boldsymbol{\mathcal{D}}^*(\gamma)\to \boldsymbol{\mathcal{D}}^*(0) ~~\mbox{as}~~\gamma\to 0.
\end{equation}
However, one can verify that $\boldsymbol{\mathcal{D}}^*(\gamma)$ is precisely $\bbeta^\gamma$, the minimizer of $\mathcal{L} (\bbeta;\gamma)$. This is because for each $\gamma\in\Gamma$:
\begin{equation}\label{ineqChain}\mathcal{L} (\bbeta^\gamma;0) \leq \mathcal{L} (\bbeta^\gamma;\gamma) \leq \mathcal{L} (\boldsymbol{0};\gamma) \leq \mathcal{L} (\boldsymbol{0};1).
\end{equation}
The first and third inequalities in \eqref{ineqChain} are thanks to $\mathcal{L} (\bbeta;\gamma)$ being increasing in $\gamma$ for a fixed $\bbeta$. 
Basically, \eqref{ineqChain} reveals that when $\gamma\in\Gamma$, $\bbeta^\gamma$  belongs to $\boldsymbol{\mathcal{D}}_1$, and $\boldsymbol{\mathcal{D}}^*(\gamma)$ has to uniquely return $\bbeta^\gamma$. Knowing that $\bbeta^\gamma$ is identical to $\boldsymbol{\mathcal{D}}^*(\gamma)$, \eqref{eqDStar}
implies that $\bbeta^\gamma$ converges to 
$\boldsymbol{\mathcal{D}}^*(0)=\bbeta_0$ as $\gamma\to 0$.\\

\noindent \textbf{Part (d).} Recall that 
\begin{align*}
 \mathcal{L}(\bbeta;\gamma) =  \sigma^2 &+ \sum_{j=1}^M \pi_j\!\left( \left\|\bbeta-\bbeta_0\right\|^2_{\bSigma - {\bar \bSigma_{\eM_j}^E}} + \left\|\bbeta_0\right\|^2_{ {\bar \bSigma_{\eM_j}^E}} + \gamma^2 \left\|\bbeta \right\|^2_{\bar \bSigma_{\eM_j}^E} \right) \\ &+    2\gamma\sqrt{\frac{2}{\pi}}\sum_{j=1}^M \pi_j \left\|\bbeta \right\|_{\bar \bSigma_{\eM_j}^E}  \left( \sigma^2 + \left\|\bbeta-\bbeta_0\right\|^2_{\bSigma - {\bar \bSigma_{\eM_j}^E}} + \left\|\bbeta_0\right\|^2_{ {\bar \bSigma_{\eM_j}^E}}\! \right)^{\frac{1}{2}}.
\end{align*}
As discussed above, $\bSigma$ being positive definite implies that $\bar\bSigma_{\eM_j}\succ \boldsymbol{0}$ for all $j\in[M]$, which explains why $\underline{\lambda}_j>0$. One can observe that the cross term in $\mathcal{L}(\bbeta;\gamma)$ can be lower bounded as follows 
\begin{align*}
    \sum_{j=1}^M \pi_j \left\|\bbeta \right\|_{\bar \bSigma_{\eM_j}^E}  \left( \sigma^2 + \left\|\bbeta-\bbeta_0\right\|^2_{\bSigma - {\bar \bSigma_{\eM_j}^E}} + \left\|\bbeta_0\right\|^2_{ {\bar \bSigma_{\eM_j}^E}}\! \right)^{\frac{1}{2}} &\geq  \sum_{j\in J} \pi_j \left\|\bbeta \right\|_{\bar \bSigma_{\eM_j}^E}  \left( \sigma^2 +  \left\|\bbeta_0\right\|^2_{ {\bar \bSigma_{\eM_j}^E}}\! \right)^{\frac{1}{2}}\\ &=  \sum_{j\in J} \pi_j \left\|\bbeta_{\eM_j} \right\|_{\bar \bSigma_{\eM_j}}  \left( \sigma^2 +  \left\|\bbeta_{0\eM_j}\right\|^2_{ {\bar \bSigma_{\eM_j}}}\! \right)^{\frac{1}{2}}\\ &\geq   \sum_{j\in J} \pi_j \underline{\lambda}_j\left\|\bbeta_{\eM_j} \right\|  \left( \sigma^2 +  \underline{\lambda}_j^2\left\|\bbeta_{0\eM_j}\right\|^2\! \right)^{\frac{1}{2}}\\& \geq  \frac{1}{\sqrt{2}}\sum_{j\in J} \pi_j \underline{\lambda}_j\left\|\bbeta_{\eM_j} \right\|  \left( \sigma +  \underline{\lambda}_j\left\|\bbeta_{0\eM_j}\right\|\! \right) \\ & \geq   \frac{1}{\sqrt{2}}\sum_{j\in J} \kappa_{\min} \left\| \bbeta_{\eM_j} \right\|\\ &\geq  \frac{\kappa_{\min}}{\sqrt{2}} \left\| \bbeta \right\|.
\end{align*}
In the above chain of inequalities, the first inequality holds because $\bSigma - {\bar \bSigma_{\eM_j}^E}\succeq \boldsymbol{0}$, the third inequality is an implication of $\sqrt{2(a^2+b^2)}\geq |a+b|$, and the last inequality is thanks to the fact that $\bigcup_{j\in J}\eM_j= [p]$. 
Now consider the function
\begin{align*}
 \tilde {\mathcal{L}}(\bbeta;\gamma) =  \sigma^2 &+ \sum_{j=1}^M \pi_j\!\left( \left\|\bbeta-\bbeta_0\right\|^2_{\bSigma - {\bar \bSigma_{\eM_j}^E}} + \left\|\bbeta_0\right\|^2_{ {\bar \bSigma_{\eM_j}^E}} + \gamma^2 \left\|\bbeta \right\|^2_{\bar \bSigma_{\eM_j}^E} \right) +  \gamma\sqrt{\frac{4}{\pi}}  \kappa_{\min} \left\| \bbeta \right\|.
\end{align*}
Clearly, for all $\bbeta$ we have $ {\mathcal{L}}(\bbeta;\gamma)\geq \tilde {\mathcal{L}}(\bbeta;\gamma)$, and specifically for $\bbeta=\boldsymbol{0}$, we get $ {\mathcal{L}}(\boldsymbol{0};\gamma) = \tilde {\mathcal{L}}(\boldsymbol{0};\gamma)$. 

To prove the theorem's statement we show that when $\gamma$ is sufficiently large, $\bbeta=\boldsymbol{0}$ is a minimizer to the convex function $\tilde{\mathcal{L}}(\bbeta; \gamma)$, which would complete the proof because then we would have
\[\forall \bbeta: ~~\mathcal{L} (\bbeta; \gamma) - \mathcal{L} (\boldsymbol{0}; \gamma)\geq \tilde{\mathcal{L}} (\bbeta; \gamma) - \tilde{\mathcal{L}} (\boldsymbol{0}; \gamma)\geq 0,
\]
making $\bbeta$ the global minimizer of $\mathcal{L} (\bbeta; \gamma)$. To proceed with showing that $\bbeta=\boldsymbol{0}$ is the minimizer of $\tilde{\mathcal{L}}(\bbeta; \gamma)$ when $\gamma$ is sufficiently large, it suffices to show that $\boldsymbol{0}\in \partial \tilde{\mathcal{L}}(\boldsymbol{0}; \gamma)$, where $\partial \tilde{\mathcal{L}}(\boldsymbol{0}; \gamma)$ is the subdifferential of $\tilde{\mathcal{L}}(\bbeta; \gamma)$ evaluated at $\bbeta=\boldsymbol{0}$. Standard subgradient calculus reveals that
\begin{equation*}
    \partial \tilde{\mathcal{L}} (\boldsymbol{0} ; \gamma) = \left\{ - 2\sum_{j=1}^M\pi_j\left(\bSigma - \bar\bSigma_{\eM_j}^E \right) \bbeta_0 +  \gamma\sqrt{\frac{4}{\pi}} \kappa_{\min}\balpha :\|\balpha\|\leq 1 \right\}.
\end{equation*}
We clearly have $\boldsymbol{0}\in \partial \tilde{\mathcal{L}}(\boldsymbol{0}; \gamma)$ when
\[\left\| \frac{\sqrt{\pi}}{\gamma\kappa_{\min}} \sum_{j=1}^M\pi_j\left(\bSigma - \bar\bSigma_{\eM_j}^E \right) \bbeta_0 \right\|\leq 1,
\]
or equivalently when
\[\gamma\geq \left\| \frac{\sqrt{\pi}}{\kappa_{\min}} \sum_{j=1}^M\pi_j\left(\bSigma - \bar\bSigma_{\eM_j}^E \right) \bbeta_0 \right\|.
\]

\subsection{Proof of Proposition \ref{propbias}}\label{prop3proof}
To simplify the notation we define the vectors $\bdelta_i\in\R^p$, with the components 
\begin{equation}\label{eqdelta}
\delta_{ij} = 1_{j\in\Ai}, ~~~i\in[n], ~j\in[p].
\end{equation}
Basically, $\bdelta_i$ is an indicator of the available components in $\vx_i$. In the derivations below we frequently use the facts that $\sum_{i=1}\delta_{ij}\delta_{ik}=np_{jk}$, and 
\[\EE x_{i,j}x_{i',k} = 1_{i=i'}\Sigma_{jk} + \mu_j\mu_k, ~~ i,i'\in [n].
\]
Verifying the unbiased nature of the mean is straightforward as
\begin{equation*}
\EE \hat \mu_j = \frac{\sum_{i=1}^n \EE x_{i,j}\delta_{ij}}{\sum_{i=1}^n \delta_{ij}} = \frac{\sum_{i=1}^n \mu_j\delta_{ij}}{\sum_{i=1}^n \delta_{ij}} =\mu_j.
\end{equation*}
To calculate the mean of the covariance entries we have
\begin{align}\notag \EE \hat\Sigma_{jk} &= \frac{\sum_{i=1}^n\delta_{ij}\delta_{ik}\EE  (x_{i,j}-\hat\mu_j)(x_{i,k}-\hat\mu_k)}{\sum_{i=1}^n\delta_{ij}\delta_{ik}} \\&= \frac{\sum_{i=1}^n\delta_{ij}\delta_{ik}\left(\EE  x_{i,j}x_{i,k}-\EE x_{i,j}\hat\mu_k -\EE\hat\mu_jx_{i,k}+\EE\hat\mu_j\hat\mu_k\right) }{np_{jk}}. \label{coveq}
\end{align}
Below we calculate each expectation in the numerator of \eqref{coveq}, separately. For the first term, we trivially have
\begin{equation}\label{covterm1}
\EE  x_{i,j}x_{i,k} = \Sigma_{jk}+\mu_j\mu_k.
\end{equation}
Expanding the second expectation term we get
\begin{align}\notag 
\EE x_{i,j}\hat\mu_k &= \frac{\EE \sum_{i'=1}^n x_{i,j} x_{i',k}\delta_{i'k}}{np_{kk}}\\\notag & = \frac{\sum_{i'=1}^n \left(1_{i=i'}\Sigma_{jk} + \mu_j\mu_k\right )\delta_{i'k} }{np_{kk}} \\ \notag &= \frac{\delta_{ik}\Sigma_{jk}+ \mu_j\mu_k \sum_{i'=1}^n \delta_{i'k} }{np_{kk}} \\&= \frac{\delta_{ik}\Sigma_{jk}}{np_{kk}}+ \mu_j\mu_k. 
\label{covterm2}
\end{align}
In a similar fashion we acquire
\begin{equation}\label{covterm3}
\EE\hat\mu_jx_{i,k} = \frac{\delta_{ij}\Sigma_{jk}}{np_{jj}}+ \mu_j\mu_k. 
\end{equation}
Last, expanding the forth term we get
\begin{align}\notag 
\EE\hat\mu_j\hat\mu_k &= \frac{1}{n^2p_{jj}p_{kk}}\sum_{i'=1}^n\sum_{i''=1}^n\delta_{i'j}\delta_{i''k}\EE x_{i',j}x_{i'',k}\\ \notag &= \frac{1}{n^2p_{jj}p_{kk}}\sum_{i'=1}^n\sum_{i''=1}^n\delta_{i'j}\delta_{i''k}\left( 1_{i''=i'}\Sigma_{jk} + \mu_j\mu_k\right) \\ \notag &=\frac{1}{n^2p_{jj}p_{kk}}\sum_{i'=1}^n \delta_{i'j}\delta_{i'k}\Sigma_{jk}+np_{kk}\mu_j\mu_k\delta_{i'j}\\ \notag &=\frac{1}{n^2p_{jj}p_{kk}}\left( np_{jk}\Sigma_{jk}+n^2p_{jj} p_{kk}\mu_j\mu_k \right) \\  &= \frac{p_{jk}}{np_{jj}p_{kk}}\Sigma_{jk}+\mu_j\mu_k.
\label{covterm4}
\end{align}
Plugging \eqref{covterm1}, \eqref{covterm2}, \eqref{covterm3}, and \eqref{covterm4} in \eqref{coveq} gives
\begin{align*}
\EE \hat\Sigma_{jk} & = \Sigma_{jk}\frac{\sum_{i=1}^n\delta_{ij}\delta_{ik}\left( 1 -\frac{\delta_{ik}}{np_{kk}} - \frac{\delta_{ij}}{np_{jj}}  + \frac{p_{jk}}{np_{jj}p_{kk}} \right) }{np_{jk}} \\& = \Sigma_{jk}\left( 1 -\frac{1}{np_{kk}} - \frac{1}{np_{jj}}  + \frac{p_{jk}}{np_{jj}p_{kk}} \right)  \frac{\sum_{i=1}^n\delta_{ij}\delta_{ik}}{np_{jk}} \\&= \Sigma_{jk}\left( 1 - \frac{p_{jj}+p_{kk}-p_{jk}}{np_{jj}p_{kk}}\right),
\end{align*}
where in the second equation we used the fact that $\delta_{ij}^2= \delta_{ij}$ and $\delta_{ik}^2= \delta_{ik}$. This completes the proof.

\subsection{Proof of Proposition \ref{propproj}}\label{prop4proof}
From \cite{halmos1972positive, higham1988matrix}, we know that since $\hat\bSigma$ is symmetric, the solution to (13) is in the form of $\hat\bSigma^+ = \hat\bSigma+ \delta \mI$, where 
\[ \delta = \min\{r: ~ \hat\bSigma+r\mI\succeq \boldsymbol{0}, ~ r\geq 0\}.
\]
On the other hand, we know that if $\lambda_1,\ldots,\lambda_p$ are the eigenvalues of $\hat\bSigma$, the eigenvalues of $\hat\bSigma+ r \mI$ are $\lambda_1+r,\ldots,\lambda_p+r$. This immediately reveals that $\delta = - \min(0,\lambda_{\min}(\hat\bSigma))$.

\subsection{Proof of Theorem \ref{thConc}}\label{th4proof}
A central tool to the proof is the Hanson-Wright inequality \cite{rudelson2013hanson} stated below.

\begin{lemma}[Hanson-Wright]\label{HWlem}
Let $\vx = (x_1,\ldots,x_p)\in \R^p$ be a random
vector with independent components $x_i$, which satisfy $\EE x_i=0$ and $\|x_i\|_{\psi_2}\leq K$ (the Orlicz-2 norm bounded by $K$). Let $\mA$ be a $p\times p$ matrix. Then, for every $t\geq 0$:
\begin{align}
\mathbb{P}\left\{\left| \vx^\top\mA\vx - \EE\vx^\top\mA\vx \right|>t\right\}  \leq 2\exp\left[ -c\min\left(\frac{t^2}{K^4\|\mA\|_F^2},\frac{t}{K^2\|\mA\|}\right)\right], \label{hwineq}
\end{align}
where $c>0$ is a universal constant. 
\end{lemma}

\textbf{Proof of part (a).} For a more compact notation, we define the variables 
\begin{equation}
d_{ij} = \frac{\delta_{ij}}{np_{jj}} ~~~i\in[n], ~j\in[p],
\end{equation}
where $\delta_{ij}$ follows the formulation in \eqref{eqdelta}. Accordingly, we define $\vd_i\in\R^p$ as a vector with entries $d_{ij}$, and $\mD_i$ as the corresponding diagonal matrix, i.e., 
\begin{equation}
\mD_i = \mbox{diag}(\vd_i)\in \R^{p\times p}.
\end{equation}
The matrix $\bSigma$ is positive semi-definite, and hence there exists $\mC\in\R^{p\times p}$, such that $\bSigma = \mC\mC^\top$,
and one can represent the vectors $\vx_i$ as $\vx_i = \mC\vz_i+\bmu$, where $\vz_i\sim \mathcal{N}(\boldsymbol{0},\mI)$.
The difference of $\bmu$ and $\hat\bmu$ can then be written as
\begin{align}\notag 
\hat\bmu - \bmu &= \sum_{i=1}^n \vd_i\odot \vx_i - \bmu = \sum_{i=1}^n \left(\vd_i\odot \vx_i - \vd_i\odot \bmu \right)\\ &= \sum_{i=1}^n \mD_i\mC\vz_i= \mD\left( \mI_n\otimes \mC\right) \vz, \label{mucent}
\end{align}
where $\mD = [\mD_1,\mD_2,\ldots, \mD_n]\in \R^{p\times np}$ and $\vz = [\vz_1;\vz_2;\ldots,\vz_n]\in\R^{np}$. In the second equality we used the fact that $\sum_{i=1}^n\vd_i = \boldsymbol{1}$, and in the third equality we used $\vd_i\odot(\vx_i-\bmu) = \mD_i(\vx_i-\bmu)$, which is a basic property of the Hadamard product. This formulation allows using the Hanson-Wright inequality, by viewing the squared norm of $\|\hat \bmu-\bmu\|^2$ as
\[\|\hat \bmu-\bmu\|^2 = \vz^\top \left( \mI_n\otimes \mC^\top \right) \mD^\top \mD\left( \mI_n\otimes \mC\right)\vz, 
\]
and considering $\mA$ in \eqref{hwineq} to be $\left( \mI_n\otimes \mC^\top \right) \mD^\top \mD\left( \mI_n\otimes \mC\right)$. In the sequel we calculate and bound the parameters used in \eqref{hwineq}. The mean of $\|\hat \bmu-\bmu\|^2$ is
\begin{align}\notag 
\EE \|\hat \bmu-\bmu\|^2 &= \EE \sum_{i=1}^n\sum_{i'=1}^n\vz_i^\top \mC^\top \mD_i^\top  \mD_{i'}\mC\vz_{i'}\\\notag & = \EE \sum_{i=1}^n\vz_i^\top \mC^\top \mD_i^\top  \mD_{i}\mC\vz_{i}
\\\notag &=\EE\sum_{i=1}^n\trace\left(  \mD_i^\top  \mD_{i}\mC\vz_{i}\vz_i^\top \mC^\top\right)\\\notag &= \sum_{i=1}^n \trace\left(  \mD_i^\top  \mD_{i}\mC\EE(\vz_{i}\vz_i^\top) \mC^\top\right)\\ &= \trace\left( \mQ\bSigma\right), \label{eqmean}
\end{align}
where $\mQ = \sum_{i=1}^n\mD_i^\top \mD_i$. The matrix $\mQ\in\R^{p\times p}$ is a diagonal matrix with the entries 
\begin{equation*}
Q_{jj} = \sum_{i=1}^n d_{ij}^2 = \frac{\sum_{i=1}^n \delta_{ij}^2}{n^2p_{jj}^2} = \frac{np_{jj}}{n^2p_{jj}^2} = \frac{1}{np_{jj}}.
\end{equation*}
Based on this observation $Q_{jj}\leq (np_{\min})^{-1}$, and we can bound the expectation as
\begin{align}\notag
\EE \|\hat \bmu-\bmu\|^2&=\trace\left( \mQ\bSigma\right) =\sum_{j=1}^p Q_{jj}\Sigma_{jj}\\&\leq \frac{1}{np_{\min}}\sum_{j=1}^p \Sigma_{jj} = \frac{1}{np_{\min}}\trace\left(\bSigma\right). \label{eqmean2}
\end{align}
Next, we bound the Frobenius and spectral norms of $\mA$. For the Frobenius norm we get
\begin{align}\notag 
\left \|\left( \mI_n\otimes \mC^\top \right) \mD^\top \mD\left( \mI_n\otimes \mC\right)\right\|_F &= \left \|\mD\left( \mI_n\otimes \mC\right) \left( \mI_n\otimes \mC^\top \right) \mD^\top \right\|_F\\\notag & = \left\| \mD\left( \mI_n\otimes \bSigma\right) \mD^\top  \right\|_F  \\\notag & = \left\| \sum_{i=1}^n \mD_i\bSigma\mD_i\right\|_F \\\notag & = \left\| \left(\sum_{i=1}^n \vd_i\vd_i^\top\right)\odot\bSigma\right\|_F \\ &\leq \frac{1}{np_{\min}}\|\bSigma\|_F.\label{eqfro}
\end{align}
In the chain of relationships above, the first equality is thanks to $\|\mB\mB^\top\|_F = \|\mB^\top \mB\|_F$ which holds for any matrix $\mB$, the forth equality is thanks to the general equation $(\vu\vv^\top)\odot\mB = \mbox{diag}(\vu)\mB\mbox{diag}(\vv)$, and the inequality is thanks to the fact that all elements of the matrix $\sum_{i=1}^n \vd_i\vd_i^\top$ are upper-bounded as 
\begin{align*}
\left[   \sum_{i=1}^n \vd_i\vd_i^\top \right]_{jk} &= \sum_{i=1}^n d_{ij}d_{ik} =\frac{\sum_{i=1}^n \delta_{ij} \delta_{ik}}{n^2p_{jj}p_{kk}}\\&\leq  \frac{\sum_{i=1}^n\delta_{ij}^2}{n^2p_{jj}p_{kk}} = \frac{1}{np_{kk}}\leq \frac{1}{np_{\min}}.
\end{align*}
The spectral norm also maintains the property $\|\mB\mB^\top\| = \|\mB^\top \mB\|$, and therefore a similar chain of relationships as above gives
\begin{align}\notag 
\left \|\left( \mI_n\otimes \mC^\top \right) \mD^\top \mD\left( \mI_n\otimes \mC\right)\right\| &= \left\| \left(\sum_{i=1}^n \vd_i\vd_i^\top\right)\odot\bSigma\right\| \\& \label{eqspec}\leq \frac{1}{np_{\min}}\|\bSigma\|.
\end{align}
Here, the inequality is thanks to Theorem P6.3.4 of \cite{rao1998matrix}, which states that for two positive semi-definite matrices $\mU$ and $\mV\in\R^{p\times p}$, with $u_{\max}$ as the maximum diagonal entry of $\mU$,
\[\lambda_j (\mU\odot\mV)\leq u_{\max}\|\mV\|, ~~j\in[p],
\]
where $\lambda_j$ denotes the $j$-th eigenvalue. Since the Orlicz-2 norm of a standard normal random variable is 1, putting together \eqref{eqmean}, \eqref{eqfro} and \eqref{eqspec}, the Hanson-Wright inequality for our problem guarantees that 
\begin{align}
\mathbb{P}\left\{\left|  \|\hat \bmu-\bmu\|^2- \trace\left( \mQ\bSigma\right) \right|>t\right\} \leq 2\exp\left[ -c\min\left(\frac{(np_{\min}t)^2}{\|\bSigma\|_F^2},\frac{np_{\min}t}{\|\bSigma\|}\right)\right].\label{hwineq2}
\end{align}
For $u\geq 0$, and $\xi_1,\xi_2> 0$, the equation $\min(u/\xi_1,u^2/\xi_2)=v$, implies that $u = \max(\xi_1v,\sqrt{\xi_2 v})$. Therefore, setting the argument of the exponential in \eqref{hwineq2} to $-c\nu$, we conclude that with probability exceeding $1 - 2\exp\left(-c\nu\right)$:
\[\left|  \|\hat \bmu-\bmu\|^2- \trace\left( \mQ\bSigma\right) \right|<  \frac{1}{np_{\min}}\max\left({\nu}\|\bSigma\|, \sqrt{{\nu}}\|\bSigma\|_F \right),
\]
which after using \eqref{eqmean2} implies the advertised inequality in the Theorem.

\textbf{Proof of part (b).}  The concentration of $\hat\bSigma$ is already derived at a minimax optimal rate in \cite{cai2016minimax}. There the authors use the Hanson-Wright inequality along with an $\epsilon$-net argument to bound the concentration of $\hat\bSigma$ around $\bSigma$.  We present the result as the following lemma. 
\begin{lemma}\label{lemcovconc}
Consider the proposed covariance estimate $\hat \bSigma$, and assume the missing pattern is independent of the $\vx_i$, $i\in [n]$. Then, for all $t>0$:
\begin{align}\label{caith}
\mathbb{P}\left\{\left\| \hat\bSigma - \bSigma \right\|<t\right\} \leq 1- \exp\left[c_1+c_2p -c_3np_{\min}\min\left(\frac{t^2}{\|\bSigma\|^2},\frac{t}{\|\bSigma\|}\right)\right], 
\end{align}	
where $c_1, c_2$ and $c_3$ are universal constants. 
\end{lemma}	
To formulate this result in the theorem's advertised form, we proceed by setting the argument of the exponential in \eqref{caith} to $-\nu$, and going through a similar line of arguments as those done for \eqref{hwineq2}, which guarantees that with probability exceeding $1-\exp(-\nu)$:
\begin{align}\notag \left\| \hat\bSigma - \bSigma \right\|&\leq \max\left(\sqrt{\frac{c_1+c_2p+\nu}{c_3np_{\min}}}, \frac{c_1+c_2p+\nu}{c_3np_{\min}}\right)\|\bSigma\|\\&=\zeta(p,np_{\min},\nu)\|\bSigma\|. \label{eqzeta}
\end{align}
On the other hand, from Proposition \ref{propproj}, 
\begin{align*}
\left\|\hat\bSigma^+ - \hat\bSigma\right\| =  \max(0,-\lambda_{\min}(\hat\bSigma)),
\end{align*}
where we simply used the fact that $\min(u,0) = -\max(0,-u)$.  Notice that 
\begin{align*}\notag 
 \|\bSigma - \hat\bSigma \| &= \lambda_{\max}(\bSigma - \hat\bSigma)  \geq \lambda_{\min}(\bSigma) + \lambda_{\max}(-\hat\bSigma) \\&= \lambda_{\min}(\bSigma) - \lambda_{\min}(\hat\bSigma),
\end{align*}
where the inequality is a direct application of the Weyl's inequality. This result combined with \eqref{eqzeta} implies that with probability exceeding $1-\exp(-\nu)$:
\begin{equation*}
-\lambda_{\min}(\hat\bSigma)\leq  \zeta(p,np_{\min},\nu)\|\bSigma\|  - \lambda_{\min}(\bSigma),
\end{equation*}
and since $\max(u,0)$ is an increasing function of $u$, we can apply it to both side of the resulting inequality, and claim that with probability exceeding $1-\exp(-\nu)$:
\begin{align}\notag
\left\|\hat\bSigma^+ - \hat\bSigma\right\|  &= \max(0,-\lambda_{\min}(\hat\bSigma))\\&\leq \max\left(0,\zeta(p,np_{\min},\nu)\|\bSigma\|  - \lambda_{\min}(\bSigma) \right).\label{eqzeta2}
\end{align}
By the triangle inequality $\|\hat\bSigma^+ - \bSigma\|\leq \|\hat\bSigma^+ - \hat \bSigma\| + \|\hat\bSigma - \bSigma\|$, and therefore after applying a  union bound (i.e., $\mathbb{P}\{A+B\}\leq \mathbb{P}\{A\} + \mathbb{P}\{B\}$) to \eqref{eqzeta} and \eqref{eqzeta2}, it follows that with probability exceeding $1 - 2\exp(-\nu)$, the advertised bound holds. 

\newpage
\section{Supplementary Material}
\subsection{Proof of Proposition \ref{prop:Li}}\label{prop1Prf}
We start by restating the formulation for $\mathcal{L}_i(\bbeta)$:
\begin{equation*}
\mathcal{L}_i(\bbeta) = \max_{|\vx_{i,\Mi}^\top \bbeta_\Mi - \bar \bmu_i^\top\bbeta_\Mi|~\leq~ \gamma \left\| \bbeta_{\mathcal{M}_i}\right\|_{\bar \bSigma_i} } \frac{1}{2} \left ( y_i - \vx_{i,\Ai}^\top \bbeta_\Ai - \vx_{i,\Mi}^\top \bbeta_\Mi\right)^2.
\end{equation*}
Since the objective inside the max is a convex function of $\vx_\Mi^\top \bbeta_\Mi$, the maximum should happen at the constraint extremes and we obtain:
\begin{equation*}
\mathcal{L}_i(\bbeta) = \max   \left\{ \mathcal{L}_i(\bbeta)^-, \mathcal{L}_i(\bbeta)^+\right\},
\end{equation*}
where 
\begin{equation*}
\mathcal{L}_i(\bbeta)^\pm  = \frac{1}{2} \left (y_i - \vx_{i,\Ai}^\top \bbeta_\Ai - \bar\bmu_i^\top\bbeta_\Mi \mp \gamma \left\| \bbeta_{\mathcal{M}_i}\right\|_{\bar \bSigma_i}\right)^2.
\end{equation*}
For arbitrary $a\in\R$ and $b\geq 0$ we have 
\begin{align*}
\max\left\{\left( a + b\right)^2, \left( a  -b \right)^2\right\}   &= a^2 + b^2 + 2\max\{ ab, -ab\} \\ &= a^2 + b^2 + 2|a|b\\ &= (|a|+b)^2,
\end{align*}
where in the second equality we used the fact $2\max\{a,b\} = a+b+|a-b|$ for arbitrary $a,b\in\R$. As a result,
\begin{equation*}
\mathcal{L}_i(\bbeta) = \frac{1}{2}\left (\left | y_i - \vx_{i,\Ai}^\top \bbeta_\Ai - \bar\bmu_i^\top\bbeta_\Mi \right | +  \gamma \left\| \bbeta_{\mathcal{M}_i}\right\|_{\bar \bSigma_i}\right)^2.
\end{equation*}
Notice that when $\gamma\geq 0$ and $\bar\bSigma_i\succeq \boldsymbol{0}$, the expression $ \left | y_i - \vx_{i,\Ai}^\top \bbeta_\Ai - \bar\bmu_i^\top\bbeta_\Mi \right | +  \gamma \left\| \bbeta_{\mathcal{M}_i}\right\|_{\bar \bSigma_i}$ is a positive quantity and convex in $\bbeta$. Next we use the following lemma which is a standard result in convex analysis. 
\begin{lemma}\label{lem:conv}
	Suppose that $f(\bbeta)$ is convex in $\bbeta$ and $f(\bbeta)\geq 0$ for all $\bbeta$. Then $f^2(\bbeta)$ is also a convex function. 
\end{lemma}
A direct application of Lemma \eqref{lem:conv} guarantees that $\mathcal{L}_i(\bbeta)$ is convex, which completes the proof. 

\subsection{Proof of Theorem \ref{th:LclosedForm}}\label{th2Prf}
The goal is calculating the expectation
\begin{align*}\EE_{\eM}\EE_{\vx}& \EE_\epsilon~ \left( \left |y - \vx_{\eM^c}^\top \bbeta_{\eM^c} - \bar\bmu_\eM^\top \bbeta_{\mathcal{M}} \right|+ \gamma \left\| \bbeta_{\mathcal{M}}\right\|_{\bar \bSigma_\eM} \right)^2 \\& =  \EE_{\eM}\EE_{\vx} \EE_\epsilon ~ \left( \left |\bbeta_0^\top\vx - \bbeta^\top\mS_{\eM^c}^\top \mS_{\eM^c} \vx - \bbeta^\top \mS_{\eM}^\top \bSigma_{\eM\eM^c } \bSigma_{\eM^c\eM^c}^{-1}\mS_{\eM^c}\vx + \epsilon\right|+  \gamma \left\| \bbeta_{\mathcal{M}}\right\|_{\bar \bSigma_\eM} \right)^2, 
\end{align*}
where $\mS_{\eM}$ follows the construction provided in (11). 

Consider $\vq\in\R^p$ to be a fixed vector, then $\vq^\top\vx+\epsilon$ is a Gaussian random variable with mean 0 and variance $\vq^\top\bSigma\vq+\sigma^2 = \|\vq\|_{\bSigma}^2 + \sigma^2$. Therefore, for this random variable we have
\begin{equation*}
    \EE \left(\vq^\top\vx+\epsilon \right)^2 = \|\vq\|_{\bSigma}^2 + \sigma^2,~~\mbox{and}~~ \EE \left|\vq^\top\vx+\epsilon\right| =\sqrt{\frac{2}{\pi}}\left( \|\vq\|_{\bSigma}^2 + \sigma^2\right)^\frac{1}{2}.
\end{equation*}
As a result
\begin{align}\notag 
      \EE_{\vx}\EE_\epsilon &\left( \left |\bbeta_0^\top\vx - \bbeta^\top\mS_{\eM^c}^\top \mS_{\eM^c} \vx - \bbeta^\top \mS_{\eM}^\top \bSigma_{\eM\eM^c } \bSigma_{\eM^c\eM^c}^{-1}\mS_{\eM^c}\vx + \epsilon\right|+  \gamma \left\| \bbeta_{\mathcal{M}}\right\|_{\bar \bSigma_\eM} \right)^2 \\ &= \sigma^2  + \left\| \vq_{\eM} \right\|^2_{\bSigma} +  \gamma^2 \left\| \bbeta_{\mathcal{M}}\right\|_{\bar \bSigma_\eM}^2 + 2\gamma \sqrt{\frac{2}{\pi}} \left\| \bbeta_{\mathcal{M}}\right\|_{\bar \bSigma_\eM} \left( \sigma^2 + \|\vq_\eM \|_{\bSigma}^2 \right) ^\frac{1}{2} ,\label{exzeta}
\end{align}
where
\[\vq_\eM = \bbeta_0 - \mS_{\eM^c}^\top \mS_{\eM^c} \bbeta - \mS_{\eM^c}^\top \bSigma_{\eM^c\eM^c}^{-1} \bSigma_{\eM^c\eM}\mS_{\eM}\bbeta .
\]
An expansion of $\|\vq_\eM \|_{\bSigma}^2$ gives
\begin{align*} &\vq_{\eM}^\top\bSigma\vq_\eM  \\&= \left(\bbeta_0 - \mS_{\!\eM^c}^\top \mS_{\!\eM^c} \bbeta - \mS_{\!\eM^c}^\top \bSigma_{\eM^c\eM^c}^{-1} \bSigma_{\eM^c\eM}\mS_{\!\eM}\bbeta \right)^\top \!\!\bSigma\! \left( \bbeta_0 - \mS_{\!\eM^c}^\top \mS_{\!\eM^c} \bbeta - \mS_{\!\eM^c}^\top \bSigma_{\eM^c\eM^c}^{-1} \bSigma_{\eM^c\eM}\mS_{\!\eM}\bbeta  \right)\\ & = 
\bbeta_0^\top\bSigma\bbeta_0\\&~~~ 
+\bbeta^\top\left( \mS_{\!\eM}^\top \bSigma_{\eM{\eM^c}}\bSigma_{{\eM^c}{\eM^c}}^{-1}\bSigma_{{\eM^c}\eM}\mS_{\!\eM}\! +\! \mS_{\!{\eM^c}}^\top \bSigma_{{\eM^c}{\eM^c}}\mS_{\!{\eM^c}} \!+\! \mS_{\!\eM^c}^\top \bSigma_{{\eM^c}\eM}\mS_{\!\eM} + \mS_{\!\eM}^\top \bSigma_{\eM{\eM^c}}\mS_{\!{\eM^c}}\right) \bbeta\\
&~~~ - 2 \bbeta_0^\top \left( \mS_{\!\eM}^\top \bSigma_{\eM{\eM^c}}\bSigma_{\!{\eM^c}{\eM^c}}^{-1}\bSigma_{{\eM^c}\eM}\mS_{\!\eM}\! +\! \mS_{\!{\eM^c}}^\top \bSigma_{{\eM^c}{\eM^c}}\mS_{\!{\eM^c}} \!+\! \mS_{\!\eM^c}^\top \bSigma_{{\eM^c}\eM}\mS_{\!\eM} \!+\! \mS_{\!\eM}^\top \bSigma_{\eM{\eM^c}}\mS_{\!{\eM^c}} \right)\bbeta\\ & = \bbeta_0^\top\bSigma\bbeta_0 + \bbeta^\top \left( \bSigma - \bar \bSigma_{\eM}^E \right) \bbeta - 2\bbeta_0^\top \left( \bSigma - \bar \bSigma_{\eM}^E \right) \bbeta\\ & =  (\bbeta-\bbeta_0)^\top \left( \bSigma - \bar \bSigma_{\eM}^E \right) (\bbeta-\bbeta_0) + \bbeta_0^\top \bar \bSigma_{\eM}^E \bbeta_0\\&= \left\| \bbeta-\bbeta_0\right\|_{\bSigma - \bar \bSigma_{\eM}^E}^2 + \left\|\bbeta_0\right\|^2_{ {\bar \bSigma_{\eM}^E}}.
\end{align*}
where in the second equality we used the fact that $\bbeta_0^\top \bSigma \mS_{\eM^c}^\top = \bbeta_0^\top\mS_{\eM^c}^\top \bSigma_{{\eM^c}{\eM^c}} + \bbeta_0^\top\mS_{\eM}^\top \bSigma_{\eM{\eM^c}}$. As a result of this we get
\begin{align*}\notag 
      \EE_{\vx}\EE_\epsilon &\left( \left |\bbeta_0^\top\vx - \bbeta^\top\mS_{\eM^c}^\top \mS_{\eM^c} \vx - \bbeta^\top \mS_{\eM}^\top \bSigma_{\eM\eM^c } \bSigma_{\eM^c\eM^c}^{-1}\mS_{\eM^c}\vx + \epsilon\right|+  \gamma \left\| \bbeta_{\mathcal{M}}\right\|_{\bar \bSigma_\eM} \right)^2 \\ = &~ \sigma^2 + \left\|\bbeta-\bbeta_0\right\|^2_{\bSigma - {\bar \bSigma_{\eM}^E}} + \left\|\bbeta_0\right\|^2_{ {\bar \bSigma_{\eM}^E}} + \gamma^2 \left\|\bbeta \right\|^2_{\bar \bSigma_{\eM}^E} \\& + 2\gamma\sqrt{\frac{2}{\pi}} \left\|\bbeta \right\|_{\bar \bSigma_{\eM}^E}  \left( \sigma^2 + \left\|\bbeta-\bbeta_0\right\|^2_{\bSigma - {\bar \bSigma_{\eM}^E}} + \left\|\bbeta_0\right\|^2_{ {\bar \bSigma_{\eM}^E}} \right)^{\frac{1}{2}}.
\end{align*}
Since $\eM$ is independent of $\vx$, and follows one of the $M$ preset patterns $\{\eM_1,\ldots,\eM_M\}$, with probability $\pi_j>0$, taking an expectation of the expression above with respect to $\eM$ gives \eqref{th:eqLBeta}.

\subsection{Proof of Proposition \ref{prop:SinglePattern}}\label{ProofOfExampleAsym}
\textbf{Part (a).} Under the proposed setting, the robust risk in \eqref{th:eqLBeta} can be written as 
\begin{align*}
    \mathcal{L} (\bbeta; \gamma) = \pi_0\sigma^2 &+\underbrace{ \pi_0\left\|\bbeta-\bbeta_0\right\|_{\bSigma}^2}_{\ell_1(\bbeta)}   +   \underbrace{\left(1-\frac{2}{\pi} \right)(1-\pi_0)\gamma^2\left\|\bbeta \right\|_{\bar\bSigma_{\eM}^E}^2}_{\ell_2(\bbeta)} \\
    & + \underbrace{(1-\pi_0) \left(  \left( \sigma^2 + \left\|\bbeta-\bbeta_0\right\|^2_{\bSigma -    \bar\bSigma_{\eM}^E  } +\left\| \bbeta_0\right\|_{\bar\bSigma_{\eM}^E }\! \right)^{\frac{1}{2}} + \sqrt{\frac{2}{\pi}}\gamma \left\|\bbeta \right\|_{\bar\bSigma_{\eM}^E} \right)^2}_{\ell_3(\bbeta)}.
\end{align*}
A similar line of argument as before guarantees the convexity of the functions $\ell_1(\bbeta)$, $\ell_2(\bbeta)$ and $\ell_3(\bbeta)$ above. Given that $\pi_0>0$, the function $\ell_1(\bbeta)$ becomes strongly (and therefore strictly) convex in $\bbeta$, and the sum of the three functions becomes strictly convex. This observation guarantees the uniqueness of the minimizer for $\mathcal{L} (\bbeta; \gamma)$. \\

\noindent \textbf{Part (b).} For a simpler notation we set $\mathcal{A}=\eM^c$, and consider the solution
\begin{equation}\label{propBeta}
    \bbeta^\gamma = \begin{bmatrix} {\bbeta_0}_{\mathcal{A}} + \bSigma_{\AAA}^{-1}\bSigma_{\AM}{\bbeta_0}_{\mathcal{M}}\\ \boldsymbol{0}  \end{bmatrix}.
\end{equation}
To prove the theorem's statement, it suffices to show that 
\begin{equation}\label{gammaCond}
\mbox{if} ~~ \gamma\geq \sqrt{\frac{\pi}{2}} \frac{\pi_0 \left\|\bbeta_0\right\|_{ \bar\bSigma_{\eM}^E} }{  (1-\pi_0)   \left( \sigma^2 + \left\|\bbeta_0\right\|^2_{\bar\bSigma_{\eM}^E}\! \right)^{\frac{1}{2}}}, ~~\forall \bbeta\in\R^p: ~~ \mathcal{L}(\bbeta; \gamma) - \mathcal{L} (\bbeta^\gamma; \gamma)\geq 0.
\end{equation}
Define the function
\begin{align}\notag 
    \tilde{\mathcal{L}} (\bbeta; \gamma) = \sigma^2 + \pi_0\left\|\bbeta-\bbeta_0\right\|_{\bSigma}^2   &+   (1-\pi_0)\left( \left\|\bbeta-\bbeta_0\right\|^2_{\bSigma - \bar\bSigma_{\eM}^E} + \left\|\bbeta_0\right\|^2_{ \bar\bSigma_{\eM}^E} + \gamma^2 \left\|\bbeta \right\|^2_{\bar\bSigma_{\eM}^E} \!\right)\\\label{eqLossSinglePatternLB}
    & +    2\sqrt{\frac{2}{\pi}} \gamma (1-\pi_0) \left\|\bbeta \right\|_{\bar\bSigma_{\eM}^E}  \left( \sigma^2 + \left\|\bbeta_0\right\|^2_{\bar\bSigma_{\eM}^E}\! \right)^{\frac{1}{2}}.
\end{align}
One can easily verify that 
\[\forall \bbeta, ~\gamma\geq 0: ~~ \tilde{\mathcal{L}} (\bbeta; \gamma)\leq \mathcal{L} (\bbeta; \gamma),
\]
and for the solution in \eqref{propBeta}: $ \tilde{\mathcal{L}} (\bbeta^\gamma; \gamma) =  \mathcal{L} (\bbeta^\gamma; \gamma)$. In the sequel, we show that when $\gamma$ meets the condition in \eqref{gammaCond}, $\bbeta^\gamma$ is also the minimizer to the convex function $\tilde{\mathcal{L}}(\bbeta; \gamma)$, which would complete the proof because then we would have
\[\forall \bbeta: ~~\mathcal{L} (\bbeta; \gamma) - \mathcal{L} (\bbeta^\gamma; \gamma)\geq \tilde{\mathcal{L}} (\bbeta; \gamma) - \tilde{\mathcal{L}} (\bbeta^\gamma; \gamma)\geq 0.
\]
To show that $\bbeta^\gamma$ is the minimizer of the convex function $\tilde{\mathcal{L}}(\bbeta; \gamma)$ when $\gamma$ meets the condition in \eqref{gammaCond}, it suffices to show that $\boldsymbol{0}\in \partial \tilde{\mathcal{L}}(\bbeta^\gamma; \gamma)$, where $\partial \tilde{\mathcal{L}}(\bbeta^\gamma; \gamma)$ is the subdifferential of $\tilde{\mathcal{L}}(\bbeta; \gamma)$ evaluated at $\bbeta^\gamma$. 

To this end, consider the function $f(\bbeta) = \|\bbeta\|_{\mS}$, where $\mS$ is a positive definite matrix. Clearly $\|.\|_{\mS}$ is a norm. It is a well-known result that (e.g., see Chapter 3 in \cite{beck2017first})
\[\partial f(\boldsymbol{0}) = \left\{\balpha: \|\balpha\|_{\mS}^* \leq 1  \right\},
\]
where $\|\cdot\|_{\mS}^*$ is the dual norm of $\|\cdot\|_{\mS}$. 
A precise characterization of the dual norm in this case is easy. Considering the Cholesky factorization $\mS=\mC\mC^\top$, we have
\begin{align*}
    \|\balpha\|_{\mS}^* = \sup_{\bsigma^\top \mS\bsigma = 1} ~\balpha^\top \bsigma = \sup_{\|\tilde\bsigma\|^2 = 1} ~\balpha^\top \mC^{-\top}\tilde \bsigma = \|\mC^{-1}\balpha\| = \|\balpha\|_{\mS^{-1}},
\end{align*}
which reveals that 
\begin{equation}\label{subgnorm}\partial f(\boldsymbol{0}) = \left\{\balpha: \|\balpha\|_{\mS^{-1}} \leq 1  \right\}.
\end{equation}
Using standard subgradient calculus and the argument above we get
\begin{equation*}
    \partial \tilde{\mathcal{L}} (\bbeta^\gamma ; \gamma) = \left\{ 2\left(\bSigma - (1-\pi_0)\bar\bSigma_{\eM}^E \right)(\bbeta^\gamma - \bbeta_0) +  2\kappa  \begin{bmatrix}\boldsymbol{0}\\ \balpha  \end{bmatrix} :\|\balpha\|_{( \bar\bSigma_{\eM})^{-1}} \leq 1 \right\},
\end{equation*}
where
\[\kappa = \sqrt{\frac{2}{\pi}} \gamma (1-\pi_0)   \left( \sigma^2 + \left\|\bbeta_0\right\|^2_{\bar\bSigma_{\eM}^E}\! \right)^{\frac{1}{2}}.
\]
Notice that
\begin{align*}
    \left(\bSigma - (1-\pi_0)\bar\bSigma_{\eM}^E \right)(\bbeta^\gamma - \bbeta_0) &= \begin{bmatrix} \bSigma_{\AAA}&\bSigma_{\AM}\\ \bSigma_{\MA} & \bSigma_{\MM}-(1-\pi_0) \bar\bSigma_{\eM}\end{bmatrix}\begin{bmatrix}  \bSigma_{\AAA}^{-1}\bSigma_{\AM}{\bbeta_0}_{\mathcal{M}}\\ -{\bbeta_0}_{\mathcal{M}}   \end{bmatrix} \\& = \begin{bmatrix}\boldsymbol{0} \\ -\pi_0 \bar\bSigma_{\eM} {\bbeta_0}_{\mathcal{M}}  \end{bmatrix},
\end{align*}
which simplifies the subdifferential to 
\begin{equation}\label{simplifiedsubg}
    \partial \tilde{\mathcal{L}} (\bbeta^\gamma ; \gamma) = \left\{   2\begin{bmatrix}\boldsymbol{0}\\- \pi_0 \bar\bSigma_{\eM} {\bbeta_0}_{\mathcal{M}} + \kappa \balpha  \end{bmatrix} :\|\balpha\|_{( \bar\bSigma_{\eM})^{-1}} \leq 1 \right\}.
\end{equation}
From \eqref{simplifiedsubg} we immediately see that $\boldsymbol{0}\in \partial \tilde{\mathcal{L}}(\bbeta^\gamma; \gamma)$ as long as
\[
\frac{\pi_0}{\kappa}\left\|  \bar\bSigma_{\eM} {\bbeta_0}_{\mathcal{M}}\right\|_{( \bar\bSigma_{\eM})^{-1}}\leq 1,
\]
or equivalently 
\[\gamma\geq  \sqrt{\frac{\pi}{2}} \frac{\pi_0 \left\|\bbeta_0\right\|_{\bar\bSigma_{\eM}^E} }{  (1-\pi_0)   \left( \sigma^2 + \left\|\bbeta_0\right\|^2_{\bar\bSigma_{\eM}^E}\! \right)^{\frac{1}{2}}}.
\]



\subsection{High Dimensional Covariance Estimation}
Section V discusses the procedure to estimate the mean and covariance in the $p<n$ regimes. When $p$ is large relative to $n$, the proposed covariance estimate is no more applicable. In these situations, one needs to consider a parametric class, or an additional structure for the covariance matrix to be able to control the  concentration. One popular parametric class is the one proposed by Bickel and Levina  \cite{bickel2008regularized}:
\begin{equation}
    \mathcal{C}(\alpha,M_0,M) = \left\{ \bSigma: \|\bSigma\|\leq M_0, ~\max_{k}\sum_j \{|\Sigma_{j,k}|: |j-k|>K\}\leq MK^{-\alpha}~\mbox{for all}~ K\right\},
\end{equation}
where a rapid decay is considered for the off-diagonal elements. Cai and Zhang \cite{cai2016minimax} propose a blockwise tridiagonal estimator of the covariance over the class $\mathcal{C}(\alpha,M_0,M)$, that is minimiax optimal, and applies to the incomplete data case. Their process still considers $\hat\bSigma$ with the earlier element estimates
\begin{align*}
\hat\Sigma_{jk}  = \frac{\sum_{i=1}^n (x_{i,j}-\hat\mu_j)(x_{i,k}-\hat\mu_k)1_{j\in\Ai}1_{k\in\Ai}}{np_{jk}}. 
\end{align*}
After picking an integer $K$, and setting $N = \lceil p/K\rceil$, they define the index sets $I_j = \{(j-1)K+1,\ldots, jK\}$ for $1\leq j\leq N-1$, and $I_N = \{(N-1)K+1,\ldots, p\}$. They then construct their estimate $\hat\bSigma^{bt}$ in a blockwise way as  
\[\hat\bSigma^{bt}_{I_j,I_{j'}} = \left\{\begin{array}{ll}\hat\bSigma_{I_j,I_{j'}}& |j-j'|\leq 1\\ \boldsymbol{0} & \mbox{otherwise} \end{array}\right., ~~~ 1\leq j,j'\leq N.
\]
They are able to show that for $K = (np_{\min})^\frac{1}{2\alpha+1}$, this simple, yet efficient estimate would satisfy the following bound conditioned on an MCAR missing pattern:
\[\sup_{\bSigma\in  \mathcal{C}(\alpha,M_0,M)} \EE \left\| \hat\bSigma^{bt} - \bSigma\right\| \leq c(np_{\min})^\frac{-2\alpha}{2\alpha+1} + c\frac{\log p}{np_{\min}},
\]
where $c$ is a constant dependent on $M$ and $M_0$. Basically, this result guarantees a reliable covariance estimate as long as $(\log p)/(np_{\min})\to 0$.

Alternatively, a popular structure considered in high dimensional covariance estimation is sparsity. In the same work  by Cai and Zhang \cite{cai2016minimax}, a procedure to estimate a sparse covariance matrix in the $(\log p)/(np_{\min})\to 0$ regime is proposed. The reader is referred to the paper for detailed discussions and results.

\subsection{More Details About the Experiments}\label{app:MoreExp}
With regards to the first set of experiments presented in Figure \ref{figSemi}, after centering and standardizing the S\&P 500 data between the years 2012 and 2017, we find out that $p=447$ companies have full data during the entire period. The sample covariance matrix obtained from these data is used to generate multi-normal data. Once the data for an experiment is generated,  we consider 100 missing patterns with the designated missing rate, and each sample has an equal chance to take any of these patterns (a MAR framework). Note that all the samples experience a missing pattern and no complete samples remain in the training data after this step. Each regression is performed 10 times, and each time the created multi-normal data takes a new set of missing patterns. 

The true parameter $\bbeta_0$ is taken to be a random vector of length $p$. To generate the response vector $\vy$, for 90\% of the samples $\vx^\top \bbeta_0$ is added a zero-mean Gaussian noise with unit variance. For the remaining 10\% of the samples, the noise has a similar nature but the standard deviation is taken to be 10. This portion basically models the outlier part of the data. 

Both for the CAM and RIGID, a free parameter needs to be tuned. This process is automated by cross validation, which obviously requires running RIGID and CAM multiple times. However, it did not seem a computational burden as both programs are convex and scalable. For $n=1000$, running RIGID takes less than 40 seconds on a 3.5 GHz Core i7 CPU, with 16 GB RAM (almost 60 ADMM iterations needed for convergence). The run time for MICE and Amelia is significantly longer, and in the fastest parameter settings at least 10 times slower than RIGID. As a general rule applied to all the experiments in the paper, the value of $\theta$ related to the estimated covariance in \eqref{eq7} is taken to be large enough to limit the condition number of $\hat\bSigma^+$ below $3\times 10^3$. Once the well-conditioning of $\hat\bSigma^+$ is guaranteed, all the conditional covariance matrices related to arbitrary missing patterns also remain well-conditioned, and no condition check is required for them. 

For the second set of experiments, and comparisons with the RT framework \cite{RosTsy09}, a similar setup as the first round of experiments was considered, with the main difference that a random covariance matrix was considered, and the true model $\bbeta_0$ was made sparse according to the designated values listed in Table \ref{RMSE_sparse}.

For the third set of experiments using real data, 10 datasets from the UCI machine learning repository are considered \cite{Dua:2019}. None of these datasets pass the multi-normality test, and the underlying hypothesis test strongly rejects their normality with p-values very close to zero. A summary of these datasets is presented in Table \ref{tab:sumUCI}.

Before running the experiments,
the data are centered and scaled, so that each column becomes zero mean with unit variance. Then, all methods are evaluated on 30 different draws of the missing masks. In each of these 30 instances, 80\% of the data are used for training, and the rest as the test set. The performance of all models are evaluated in terms of the RMSE and MAE.

\begin{table}[h]\label{tbl:data}
\caption{Summary of the UCI datasets used for comparison}
    \label{tab:sumUCI}
    \centering
    \begin{tabular}{l c c c}
    \hline
        Dataset & $n$ & $p$ & Reference \\ \hline
        QSAR aquatic toxicity & 546 & 9 & \cite{cassotti2014prediction}\\
         QSAR fish toxicity & 908 & 7 &  \cite{cassotti2015similarity}\\
         Bias correction in temperature forecast & 7750 & 25 & \cite{cho2020comparative}\\
         Combined cycle power plant  & 9568 & 4 & \cite{tufekci2014prediction}\\
         Concrete slump test & 103 & 10 & \cite{yeh2007modeling}\\
         Concrete compressive strength & 1030 & 9 & \cite{yeh1998modeling}\\
        Gas turbine CO and NOx emission & 36733 & 11 & \cite{kaya2019predicting}\\
        Wine quality (red) & 1599 & 10 & \cite{cortez2009modeling}\\
        Wine quality (white) & 4898 & 11 & \cite{cortez2009modeling} \\
        Yacht hydrodynamics & 308 & 7 & \cite{Dua:2019}\\
         \hline
    \end{tabular}
    
\end{table}

In terms of the missing mechanisms used,  we consider all the three standard mechanisms, namely, MCAR, MAR and MNAR. In the case of MNAR, logistic and quantile-based masks (MNAR\_q) are used.
We refer the reader to \cite{muzellec2020missing} and their referenced GitHub page for a detailed description of these mechanisms and the 
implementation in Python. A more detailed summary about the application of each mechanism is presented below:

\begin{enumerate}
    \item MCAR  50\%: The value of each variable in each observation is masked independently with probability  0.5.
    
    \item MAR 50\%: In this case,  30\%  of the variables with no missing entries are randomly selected. The remaining variables take missing patterns according to a logistic model with random weights,  re-scaled to attain the desired missing rate of 50\%.
    
    \item MNAR 50\%: In this case, the variables are first split into a set of inputs for a logistic model, and a set whose missing probabilities are determined by a logistic model. We use a 30\% randomly selected potion of the variables as the input of the logistic model.
    Then inputs are masked according to an MCAR mechanism. The logistic model weights are randomly selected, and the intercept term is adjusted to attain the desired missing rate of 50\%.
    
    \item MNAR\_q 50\%:
    First, 70\% of the variables with missing entries are randomly selected. Then, only applied to the upper or lower 25\% quantiles, the values are masked with probability 0.5. Since the missingness depends on the quantile information and hence the masked values, the framework follows an MNAR mechanism.
\end{enumerate}

\begin{figure}[!htbp]
\centering\begin{overpic}[trim={0 .7cm  0 0},clip,width=3in]{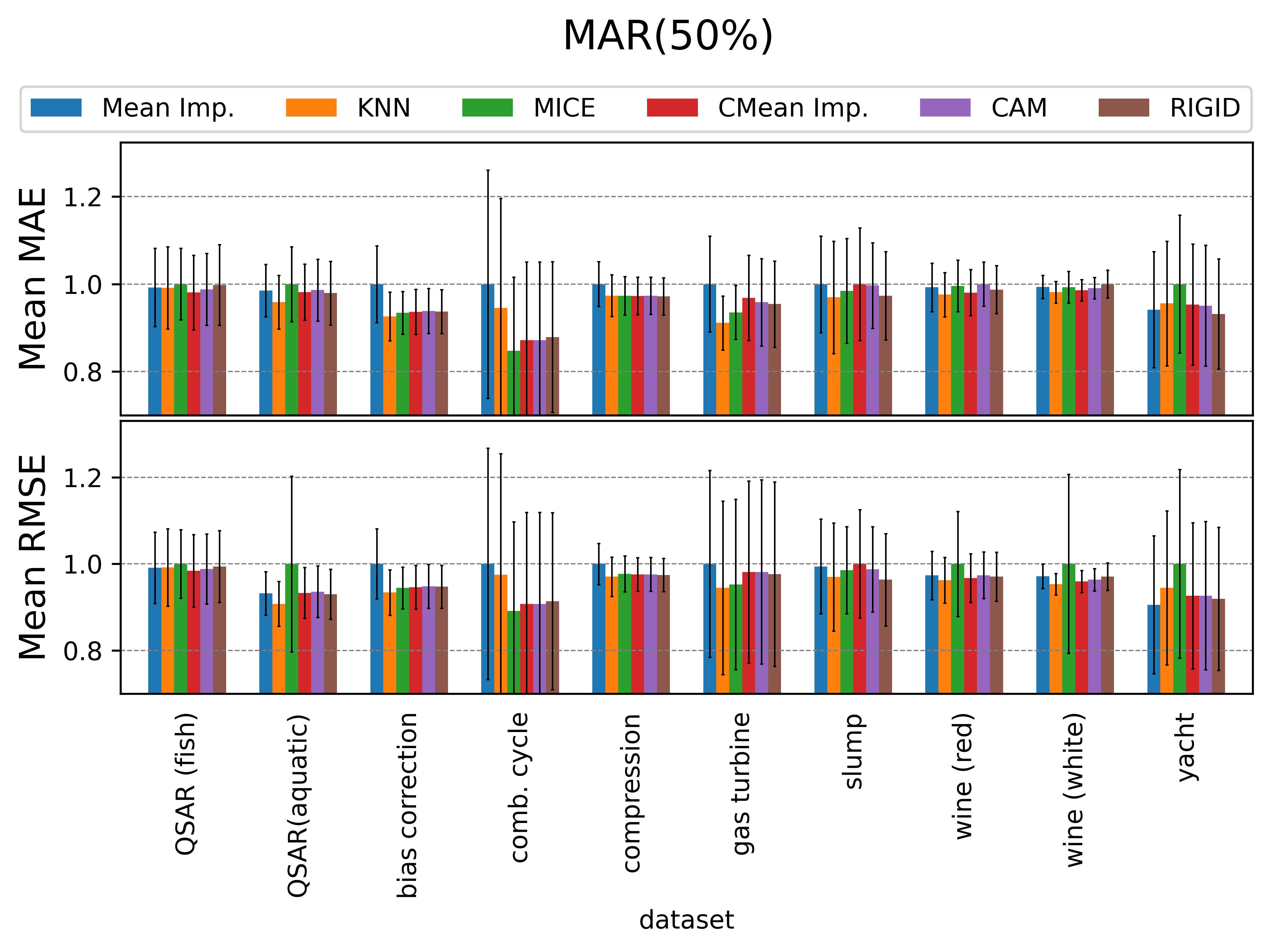}
\end{overpic}
\begin{overpic}[trim={0 .7cm  0 0},clip,width=3in]{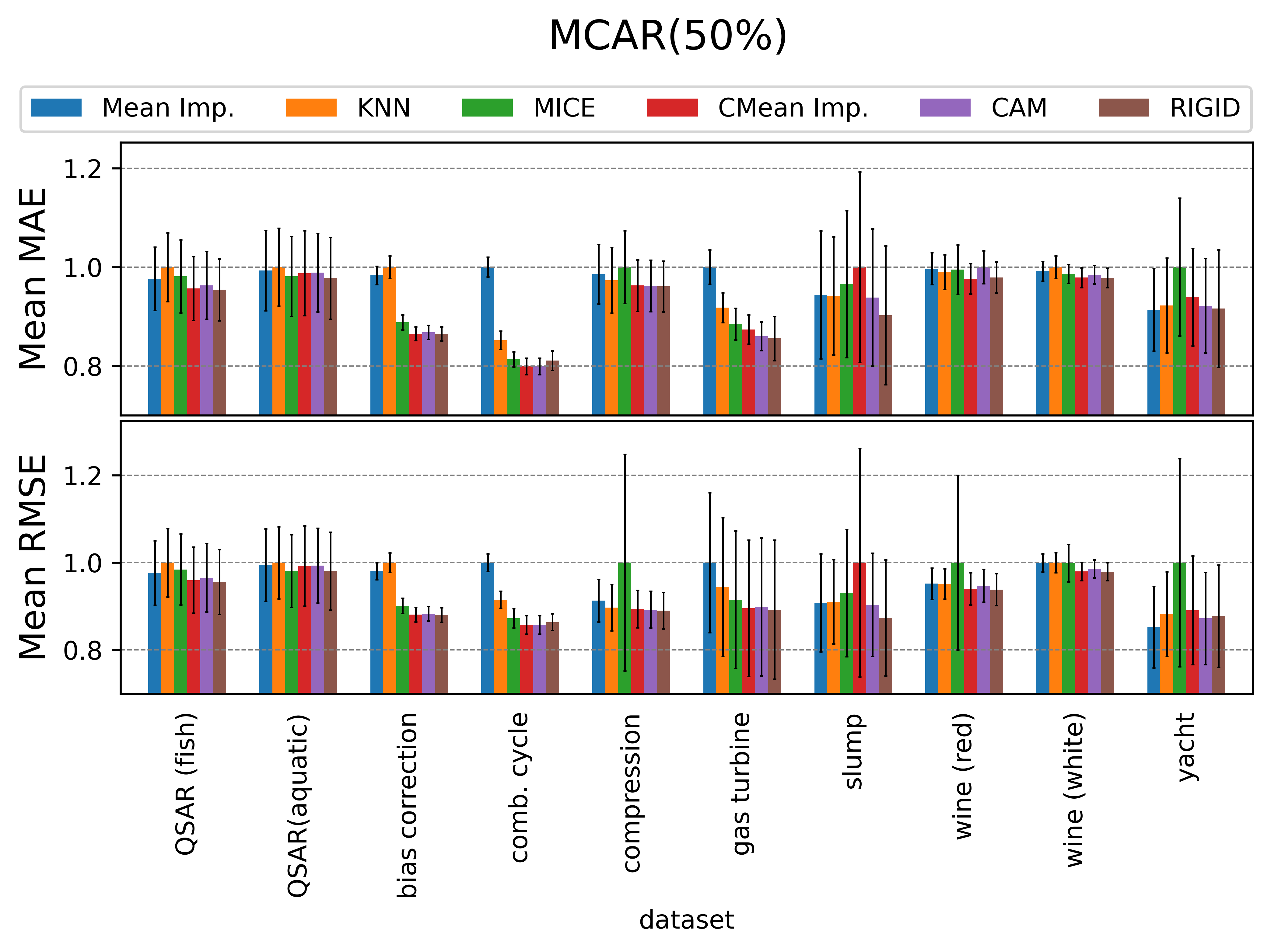}
\end{overpic}
\begin{overpic}[trim={0 .7cm  0 0},clip,width=3in]{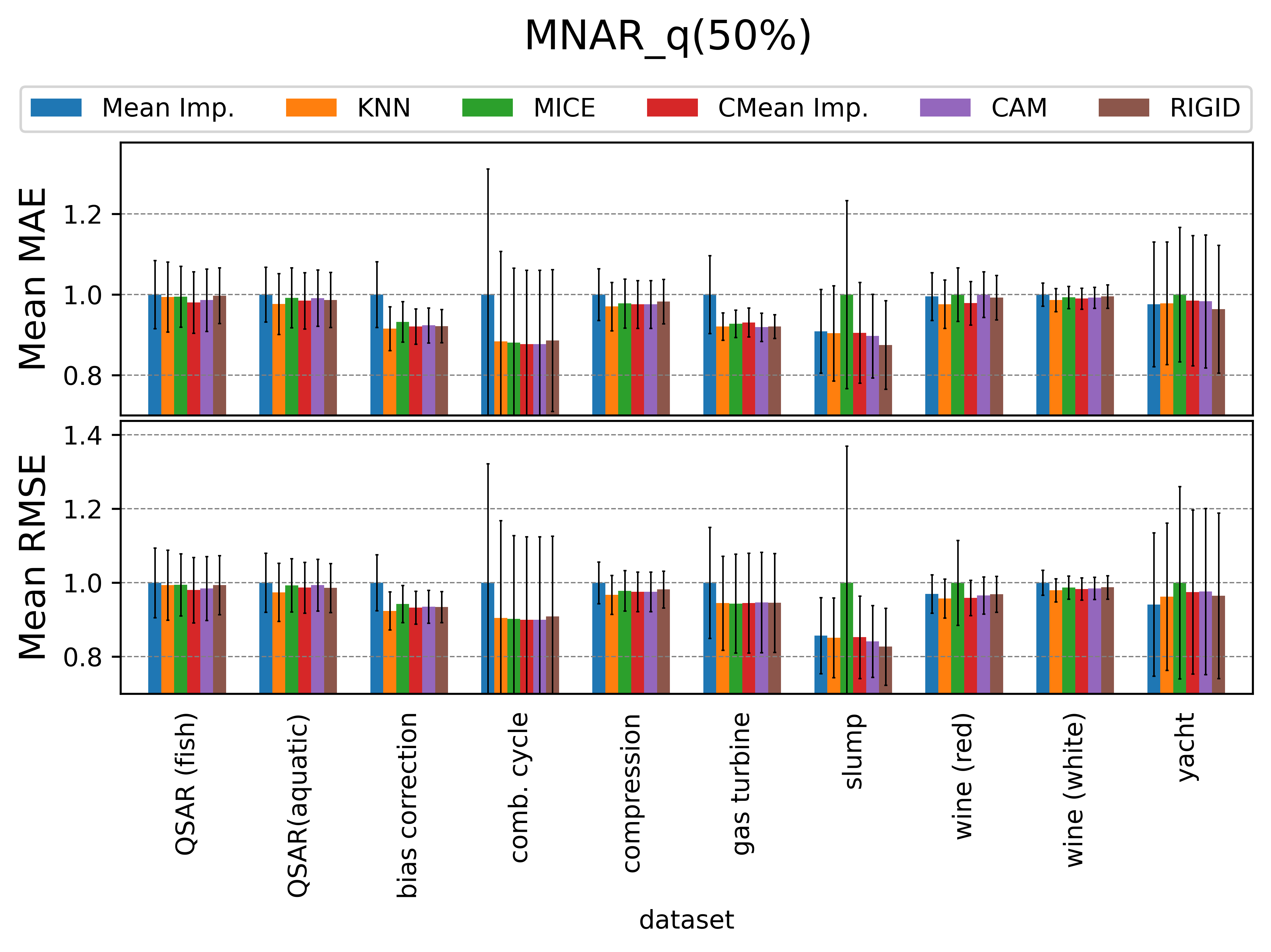}
\end{overpic}
\begin{overpic}[trim={0 .7cm  0 0},clip,width=3in]{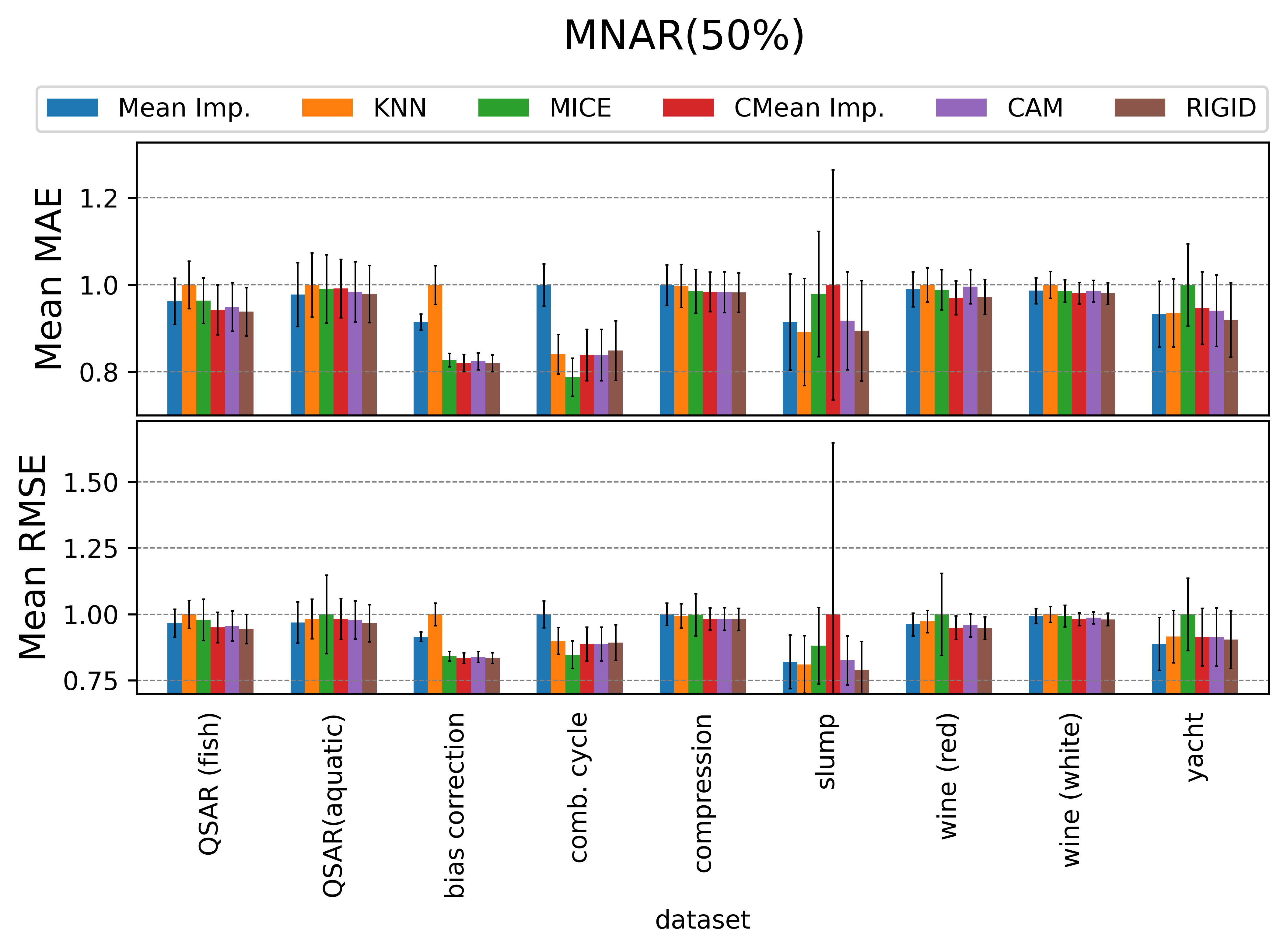}
\end{overpic}\vspace{-.3cm}
\caption{More experiments related to other missing mechanisms }\label{figMoreExp}
\end{figure}
Figure \ref{figMoreExp} presents the average MAE and RMSE values for an extensive set of experiments following the aforementioned missing mechanisms.  The error bars correspond to one standard deviation of the MAE/RMSE values. For a better readability, the evaluations for each dataset are normalized to the maximum mean MAE (or mean RMSE). It is important to note that a linear model is intrinsically a limited model, and there is a fundamental limit on the extent the prediction error can be reduced. Figure \ref{figMoreExp} still reports RIGID as a top performing framework over the majority of the experiments. It is interesting to note that thanks to the robust formulation of RIGID, the MAE values are also notably reduced.

\end{document}